\documentclass{article}
\usepackage[accepted]{icml2021}    %

\usepackage{float}
\usepackage{wrapfig}
\usepackage{microtype}
\usepackage{graphicx}
\usepackage{subcaption}
\usepackage{booktabs} %
\usepackage[algo2e,algoruled,ruled,boxed,lined]{algorithm2e}

\usepackage{hyperref}
\definecolor{mydarkblue}{rgb}{0,0.08,0.45}  %
\hypersetup{colorlinks=true,linkcolor=red,citecolor=mydarkblue,urlcolor=magenta,menucolor=red}

\usepackage[utf8]{inputenc} %
\usepackage[T1]{fontenc}    %
\usepackage{url}            %
\usepackage{booktabs}       %
\usepackage{makecell}
\usepackage{multirow}
\usepackage{amsfonts}       %
\usepackage{nicefrac}       %
\usepackage{microtype}      %
\usepackage{amsmath,amssymb,amsthm}
\usepackage{enumitem}
\usepackage{stackengine}
\usepackage{comment}
\usepackage{lipsum}
\usepackage{placeins}

\usepackage{color}
\usepackage{xspace}
\PassOptionsToPackage{usenames,dvipsnames}{xcolor}
\usepackage{tcolorbox}
\usepackage{cleveref}
\crefname{equation}{Eq.}{}

\usepackage[normalem]{ulem}   %
\usepackage{pifont}
\newcommand{\cmark}{\ding{52}\xspace}%
\newcommand{\cmarkgreen}{{\color{green!50!black}\cmark}\xspace}%
\newcommand{\argmax}{\operatornamewithlimits{argmax}}

\DeclareRobustCommand\onedot{\futurelet\@let@token\@onedot}
\def\eg{{e.g.}\xspace} 
\def\ie{{i.e.}\xspace}

\newcommand{\R}{\mathbb{R}}
\renewcommand{\S}{\mathcal{S}}
\newcommand{\G}{\mathcal{G}}

\usepackage[colorinlistoftodos,textsize=small]{todonotes}

\usepackage{marginnote}

\providecommand{\Yes}{\hspace{3pt}\cmarkgreen\hspace*{2pt}}
\newcolumntype{P}[1]{>{\centering\arraybackslash}p{#1}}
\newcolumntype{H}{@{}>{\lrbox0}l<{\endlrbox}}

\usepackage[compact]{titlesec}
\titlespacing*{\paragraph}{0pt}{-1pt}{7pt}   %
\usepackage{lineno}
\newcommand*\patchAmsMathEnvironmentForLineno[1]{%
  \expandafter\let\csname old#1\expandafter\endcsname\csname #1\endcsname
  \expandafter\let\csname oldend#1\expandafter\endcsname\csname end#1\endcsname
  \renewenvironment{#1}%
                   {\linenomath\csname old#1\endcsname}%
                   {\csname oldend#1\endcsname\endlinenomath}%
}%
\newcommand*\patchBothAmsMathEnvironmentsForLineno[1]{%
  \patchAmsMathEnvironmentForLineno{#1}%
  \patchAmsMathEnvironmentForLineno{#1*}%
}%
\patchBothAmsMathEnvironmentsForLineno{equation}%
\patchBothAmsMathEnvironmentsForLineno{align}%
\patchBothAmsMathEnvironmentsForLineno{flalign}%
\patchBothAmsMathEnvironmentsForLineno{alignat}%
\patchBothAmsMathEnvironmentsForLineno{gather}%
\patchBothAmsMathEnvironmentsForLineno{multline}%
\def\thepapertitle{%
Variational Empowerment as Representation Learning \\for Goal-Based Reinforcement Learning
}
\title{\thepapertitle}

\icmltitlerunning{Variational Empowerment as Representation Learning for Goal-Based RL}

\begin{document}

\newcommand{\adagger}{\,\textsuperscript{\textdagger}\hspace*{-0.1em}}

\twocolumn[
\icmltitle{\thepapertitle}
\icmlsetsymbol{equal}{*}

\begin{icmlauthorlist}
\icmlauthor{Jongwook Choi\adagger}{umich}
\icmlauthor{Archit Sharma\adagger}{stanford}
\icmlauthor{Honglak Lee}{umich,lg}
\icmlauthor{Sergey Levine}{google,berkeley}
\icmlauthor{Shixiang Shane Gu}{google}
\end{icmlauthorlist}

\icmlcorrespondingauthor{Jongwook Choi}{jwook@umich.edu}
\icmlcorrespondingauthor{Shixiang Shane Gu}{shanegu@google.com}

\icmlaffiliation{google}{Google Research}
\icmlaffiliation{umich}{University of Michigan}
\icmlaffiliation{berkeley}{University of California, Berkeley}
\icmlaffiliation{stanford}{Stanford University}
\icmlaffiliation{lg}{LG AI Research}
\icmlkeywords{Reinforcement Learning}
\vskip 0.3in
]

\printAffiliationsAndNotice{%
\textsuperscript{\textdagger}Work done while an intern at Google.

} %

\vspace*{-13pt}
\begin{abstract}
    Learning to reach goal states and
    learning diverse skills through mutual information (MI) maximization have been proposed as principled frameworks for self-supervised reinforcement learning,
    allowing agents to acquire broadly applicable multi-task policies with minimal reward engineering.
    Starting from a simple observation that the standard goal-conditioned RL (GCRL) is encapsulated by the optimization objective of variational empowerment, we discuss how GCRL and MI-based RL can be generalized into a single family of methods, which we name \textit{variational GCRL (VGCRL)},
    interpreting variational MI maximization, or variational empowerment, as representation learning methods that acquire functionally-aware state representations for goal reaching.
    This novel perspective allows us to: (1) derive simple but unexplored variants of GCRL to study how adding small representation capacity can already expand its capabilities; (2) investigate how discriminator function capacity and smoothness determine the quality of discovered skills, or latent goals, through modifying latent dimensionality and applying spectral normalization; 
    (3) adapt techniques such as hindsight experience replay (HER) from GCRL to MI-based RL;
    and lastly, (4) propose 
    a novel evaluation metric, named latent goal reaching (LGR), for comparing empowerment algorithms with different choices of latent dimensionality and discriminator parameterization.
    Through principled mathematical derivations and careful experimental studies,
    our work lays a novel foundation from which to evaluate, analyze, and develop representation learning techniques in goal-based RL.
\end{abstract}

\section{Introduction}
\label{sec:introduction}
\vspace*{-4pt}

Reinforcement learning (RL) provides a general framework for discovering optimal behaviors for sequential decision-making.
Combined with powerful function approximators like neural networks, RL can be used to learn to play computer games from raw pixels~\citep{mnih2013playing} and acquire complex sensorimotor skills with real-world robots~\citep{gu2017deep,kalashnikov2018qt,haarnoja2018soft}.
Neural networks show best performance, generalization, and reusability when they are trained on large and diverse datasets~\citep{krizhevsky2012imagenet,devlin2018bert}.
However, a critical limitation in RL is that human experts often need to spend considerable efforts designing and fine-tuning reward functions per task, making it hard to scale and define a huge set of tasks in advance.
If we have agents that can interact with the world without rewards, build up a body of knowledge autonomously, and utilize this knowledge to accomplish new tasks efficiently, then we can greatly scale up task and skill learning to achieve similar level of generalization and performance for RL as what neural networks have enabled for other domains.
Several works have tried to find a single generalizable task-agnostic reward function which can potentially be used across several environments.
The use of such intrinsic reward functions has been motivated
as exploration heuristics such as curiosity and novelty~\citep{schmidhuber1991curious,oudeyer2009intrinsic, bellemare2016unifying,pathak2017curiosity},
as optimizing mutual information (MI)~\citep{Gregor:ICLRWorkshop2017:VIC,Eysenbach:ICLR2019:DIAYN, Sharma:ICLR2020:DADS, Sharma:RSS2020:off-DADS}
or as empowerment~\citep{klyubin2005all,jung2011empowerment,mohamed2015variational}.
Classically, goal-conditioned RL (GCRL) has shown success in learning diverse and useful skills in concurrence to MI-based methods.
GCRL optimizes a stationary and interpretable reward for goal-reaching, but when the goal space is high-dimensional, how does the agent know which part of the space is relevant and which part can be ignored?
In such cases, prior GCRL works frequently rely on manual definition~\citep{Andrychowicz:NIPS2017:HER} or off-the-shelf representation learning~\citep{nachum2018near,nair2018visual,wu2018laplacian} optimized prior to or separately from reinforcement learning.
Meanwhile, MI or empowerment-based RL offers a clear objective for representation learning \textit{through} reinforcement learning, but the properties of the learned behaviors are often unclear due to lack of a proper evaluation metric.
Prior works use qualitative inspections of learned behaviors, variational bound estimates, or downstream task performances of a skill-utilizing high-level policy~\citep{Eysenbach:ICLR2019:DIAYN, Sharma:ICLR2020:DADS}, but these heuristics are costly or indirect measures and make objective comparisons and analyses of various mathematically-similar MI-based algorithms difficult~\citep{florensa2017stochastic,Eysenbach:ICLR2019:DIAYN,Achiam:2018:VALOR,WardeFarley:ICLR2019:DISCERN,Hansen:ICLR2020:VISR,Sharma:ICLR2020:DADS}.
To recover a more direct metric, an important question is: what do these MI-based objectives learn representations for?
In this work, we interpret MI and empowerment-based RL as a principled framework for representation learning in goal-conditioned RL.
Starting from a simple observation that the objective of the standard GCRL
  can be seen as a special case of variational MI with a fixed hard-coded variational posterior,
our analysis provides a unification of these ideas and explicitly reframes skill discovery via mutual information maximization~\citep{Gregor:ICLRWorkshop2017:VIC,Eysenbach:ICLR2019:DIAYN} as a combination of representation learning and goal-conditioned reinforcement learning, where both the space of goals and the skills to reach those goals are learned jointly via a MI-based objective.
\iftrue          %
While the connections between representation learning, mutual information estimation, and goal-conditioned RL have been explored in a number of previous works~\citep{Gregor:ICLRWorkshop2017:VIC,WardeFarley:ICLR2019:DISCERN,gupta2018unsupervised}, our exact mathematical formulation and granular analyses enable new perspectives and synergies between GCRL and MI-based RL: 
\begin{enumerate}[leftmargin=*]
    \setlength{\itemsep}{2pt}\setlength{\parskip}{2pt}
    
    \item \textbf{[MI to GCRL]} 
    We propose simple but novel variants of GCRL -- adaptive-variance and linear-mapping GCRL -- to study how adding small representation capacity can already expand the capabilities of GCRL.
    \item \textbf{[MI to GCRL]} We show that a proper representation regularization from generative modeling, such as spectral normalization~\citep{miyato2018spectral}, can improve the quality of latent goals discovered (and the stability of MI-based algorithms).
    \item \textbf{[GCRL to MI]} We adapt hindsight experience replay (HER)~\citep{Andrychowicz:NIPS2017:HER} from GCRL to more general MI-based objectives
        and show posterior HER (P-HER) consistently provides substantial performance gains in MI-based RL algorithms.
    \item \textbf{[GCRL to MI]} We propose the latent goal reaching (LGR) metric as an intuitive, task-oriented, and discriminator-agnostic metric for objectively evaluating empowerment algorithms.
\end{enumerate}
\fi

\newpage

\section{Related Work}
\label{sec:related-work}
\vspace*{-5pt}

Reward engineering has been a bottleneck to broad application of RL. %
Some of the prior attempts to alleviate this problem have sought introduce human supervision in alternative, easier forms,
such as demonstrations~\citep{ng2000algorithms,abbeel2004apprenticeship,ziebart2008maximum,ho2016generative,fu2017learning,ghasemipour2019divergence}
or preferences~\citep{hadfield2017inverse,christiano2017deep}.
However, since these methods still rely on non-negligible amounts of human interventions, they cannot automatically scale to solving thousands of new environments and tasks.

\paragraph{Empowerment and reward-free RL.} Task-agnostic reward functions have been proposed to encourage exploration in environments using notions of curiosity or novelty~\citep{schmidhuber1991curious,oudeyer2009intrinsic,schmidhuber2010formal,bellemare2016unifying,pathak2017curiosity,colas2018curious}. In a similar vein, some methods maximize the state-visitation entropy~\citep{hazan2018provably,pong2019skew,lee2019efficient,ghasemipour2019divergence}.
These approaches can enable solutions to otherwise hard exploration sparse-reward problems. Some of the recent work has emphasized on empowerment or option/skill discovery through optimization of mutual information based intrinsic reward functions.
Classically, empowerment measures the ability of an agent to control the environment~\citep{Salge:2014:Empowerment, klyubin2005all, jung2011empowerment}, which was scaled up by~\cite{mohamed2015variational,karl2017unsupervised}.
The concept of mutual information, which is also at the heart of empowerment based methods, has been further used to motivate several objectives for skill discovery~\citep{florensa2017stochastic,Eysenbach:ICLR2019:DIAYN,Achiam:2018:VALOR,WardeFarley:ICLR2019:DISCERN,Hansen:ICLR2020:VISR,Sharma:ICLR2020:DADS}. 
Recent works have shown that skills learned through mutual information can be meaningfully combined to solve downstream tasks~\citep{Eysenbach:ICLR2019:DIAYN,Sharma:ICLR2020:DADS}, even on real robots~\citep{Sharma:RSS2020:off-DADS}.

\paragraph{Goal-conditioned RL.} Goal-conditioned RL~\citep{kaelbling1993learning,pong2018temporal,Andrychowicz:NIPS2017:HER,Schaul:ICML2015:UVFA}
provides a framework for enabling agents to reach user-specified goal states. %
The behaviors learned via GCRL can be interpretable and easy to analyze in terms of the goal-reaching function.
However, GCRL assumes that a goal-reaching function has been specified in addition to goal states, which precludes broader application for the same reasons as those for reward engineering.
Some prior works have used mutual information in the GCRL framework ~\citep{pong2019skew, WardeFarley:ICLR2019:DISCERN}.
However, these works use the MI optimization as an unsupervised scheme to generate goals.
On the other hand, our work studies skill-discovery/empowerment methods and provides an explicit reinterpretation within the GCRL framework,
combining the representation learning perspective with the goal-reaching behavior of GCRL.
\section{Background}
\label{sec:background}
\vspace*{-3pt}

In this section, we briefly review mutual information (MI)-based objectives for skill discovery,
focusing on variational approaches introduced in~\citep{mohamed2015variational, Gregor:ICLRWorkshop2017:VIC, Eysenbach:ICLR2019:DIAYN,Sharma:ICLR2020:DADS}, and goal-conditioned RL (GCRL).

We denote a Markov decision process (MDP) $\mathcal{M} = (\mathcal{S}, \mathcal{A}, p, r)$, where $\mathcal{S}$ denotes the state space, $\mathcal{A}$ denotes the action space,
$p : \mathcal{S} \times \mathcal{S} \times \mathcal{A} \rightarrow [0, \infty)$ denotes the underlying (possibly stochastic) transition dynamics of the environment with the initial state distribution $p_0: \mathcal{S} \rightarrow [0, \infty)$,
and a reward function $r: \mathcal{S} \times \mathcal{A} \rightarrow \mathbb{R}$.
The goal of the RL optimization problem is to learn a policy $\pi(a \mid s)$ which maximizes the return
$
    \mathbb{E}_{p, \pi}\left[\sum_{t=0}^\infty \gamma^tr(s_t, a_t)\right]
    = \mathbb{E}_{s\sim \rho^\pi, a\sim \pi}\left[ r(s,a) \right],
$
for a discount factor $\gamma \in [0, 1)$ where $\rho^\pi$ is an unnormalized $\gamma$-discounted state visitation density.
Importantly, once we write an objective in the form of $\mathbb{E}_{\rho^\pi, \pi}\left[ r(s,a) \right]$,
we can apply the policy gradient theorem~\citep{sutton2000policy} to derive a practical RL solver,
as done in~\citep{kakade2002natural,silver2014deterministic,schulman2015trust,gu2017interpolated,ciosek2018expected},
or learn it with Q-learning~\citep{watkins1992q}.
For simplicity of our notations, we omit the discount factor $\gamma$ in the following sections and derivations.

\vspace*{-1pt}
\subsection{Mutual Information Maximization and Empowerment}
\label{sec:bg_mi}
\vspace*{-5pt}

MI maximization in RL such as empowerment generally means maximizing the mutual information between (some representations of) actions and (some representations of) future states following those actions~\citep{klyubin2005all,mohamed2015variational}.
The goal is to learn a set of actions that can influence future states to be diverse,
but also be predictable if we know what action is taken. 
In this work we focus on learning abstract representation $z\in \mathcal{Z}$, which is an additional input to the policy $\pi(a|s,z)$ and defines a set of empowered actions.
The latent code $z$ (either discrete or continuous) can be interpreted as a macro-action,
\emph{skill} or \emph{goal}~\citep{Eysenbach:ICLR2019:DIAYN,Sharma:ICLR2020:DADS}.

We discuss two variants of MI objectives in RL:
\emph{state-predictive MI}~\citep{Sharma:ICLR2020:DADS}, which maximizes
$
    \mathcal{I}(s'; z\mid s)
$,
and
\emph{state-marginal MI}~\citep{Eysenbach:ICLR2019:DIAYN}, which maximizes
$
    \mathcal{I}(s; z)
$.
Due to page limit, we discuss these variants more in detail in \Cref{sec:extended-background}.
In this work, we focus on state-marginal MI, whose optimization objective is:
\begin{align}
    \mathcal{I}(s; z) &= \mathbb{E}_{z \sim p(z), s \sim \rho^\pi(s|z)} [ \log p(z \mid s) - \log p(z) ]  \nonumber \\
     & \geq \mathbb{E}_{z \sim p(z), s \sim \rho^\pi(s|z)} [ \log q_\lambda (z \mid s) - \log p(z) ]
     \label{eq:var_emp}
\end{align}
where $q_\lambda(z | s)$ is a variational approximation to the intractable posterior $p(z | s)$,
often called a (skill) discriminator~\citep{Eysenbach:ICLR2019:DIAYN}.

Given a parameterized policy $\pi_\theta(a|s,z)$, Eq.~\ref{eq:var_emp} gives a joint maximization objective (a variational lower bound) with respect to $\pi_\theta$ and $q_\lambda$:
\begin{align}
    \mathcal{F} (\theta, \lambda) = \mathbb{E}_{z,s\sim \pi_\theta}\left[ \log q_\lambda(z|s) - \log p(z) \right] \label{eq:joint_op}.
\end{align}
A simple iterative RL procedure can be derived to optimize this lower bound, assuming a parameterized policy $\pi_\theta(a|s,z)$, where at iteration $i$,
\vspace*{-3pt}
\begin{align}
    \lambda^{(i)} \leftarrow \textstyle\argmax_\lambda \mathbb{E}_{z,s\sim \pi^{(i-1)}} \left[ \log q_\lambda(z|s) - \log p(z) \right] \label{eq:lambda_op} \\[-2pt]
    \theta^{(i)}  \leftarrow \textstyle\argmax_\theta  \mathbb{E}_{z,s\sim \pi_\theta}  \left[ \log q_{\lambda^{(i)}}(z|s) - \log p(z) \right] \label{eq:pi_op}.
\end{align}
Eq.~\ref{eq:lambda_op} is a simple supervised regression
(\eg, maximum likelihood) on on-policy samples.
Eq.~\ref{eq:pi_op} has the same form of standard RL, and therefore can be optimized using any RL algorithm~\citep{Gregor:ICLRWorkshop2017:VIC,Eysenbach:ICLR2019:DIAYN}.

\vspace*{-1pt}
\subsection{Goal-Conditioned RL}
\label{sec:bg_gcrl}
\vspace*{-5pt}

Goal-conditioned RL~\citep{kaelbling1993learning,Schaul:ICML2015:UVFA} (GCRL) is a standard, stationary-reward problem
where we aim to find a policy $\pi(a|s,g)$ conditional on a goal $g\in\mathcal{G}$ by maximizing
\vspace*{-1pt}
\begin{align}
    F(\pi) = \mathbb{E}_{g \sim p(g), s \sim \pi_\theta}\left[ -d(s,g) \right] \label{eq:goal_op},
\end{align}
where $p(g,s)=p(g)\rho^\pi(s|g)$, $p(g)$ defines the task distribution over goals,
and $d(s,g)$ is a distance metric between state $s$ and goal $g$, such as an Euclidean distance. %
The main challenges for goal-conditioned RL lie in defining the goal space and the goal-reaching reward function $-d(s, g)$,
which often requires task-specific knowledge or careful choices of goal space~\citep{plappert2018multi}.
In off-policy learning, hindsight experience replay (HER)~\citep{kaelbling1993learning,Andrychowicz:NIPS2017:HER} has shown to improve learning of goal-conditioned policy significantly.
The key insight is that for a given exploration episode $\{g, s_{0:T}\}$, one can relabel the goal  %
with an \emph{actually} achieved goal $\mathbb{S}(s_{0:T})$, derived by a strategy function $\mathbb{S}(\cdot)$.
A typical choice is to relabel the goal as $\tilde{g}=\mathbb{S}(s_{0:T})=s_T$,
which can be seen as self-supervised curriculum learning~\citep{Andrychowicz:NIPS2017:HER,lynch2019learning}.

\vspace*{-2pt}
\section{Expressivity Tradeoffs in Variational Empowerment}
\label{sec:vge}
\vspace*{-5pt}
Interestingly, the simple objective in Eq.~\ref{eq:joint_op},
which we term \textit{Variational Goal-Conditioned RL} (VGCRL),
encapsulates most of the prior MI-based algorithms~\citep{Eysenbach:ICLR2019:DIAYN,WardeFarley:ICLR2019:DISCERN,Hansen:ICLR2020:VISR}
with the only differences being goal space $\mathcal{Z}$, prior $p(z)$, and discriminator $q_\lambda(z|s)$, as detailed in Table~\ref{tab:summary-algorithmsv2}.
For example, when $z$ is a discrete variable, this reduces to DIAYN~\citep{Eysenbach:ICLR2019:DIAYN} or VALOR~\citep{Achiam:2018:VALOR}.

\begin{table*}[t]
\small
\setlength{\tabcolsep}{3pt}
\centering
\hspace*{-11pt}
\begin{tabular}{c|cccc} %
    \toprule
    \textbf{Method}                              & Goal space          & $q_\lambda(z | s)$                   & {Learnable $\lambda$}       & Learning $\pi_z$          \\ \midrule
    GCRL~\citep{kaelbling1993learning}           & Continuous ($\R^d$) & $\mathcal{N}(s, \sigma^2 I)$         & -                           & (HER)                       \\
    aGCRL \textbf{(ours)}           & Continuous ($\R^d$) & $\mathcal{N}(s, \Sigma)$         & $\Sigma$                           & HER                       \\
    linGCRL \textbf{(ours)}                         & Continuous ($\R^d$) & $\mathcal{N}(As, \sigma^2 I)$        & $A$                         & P-HER                      \\ %
    InfoGAIL$^*$~\citep{Li:2017:InfoGAIL}        & Discrete            & Categorical                  & $q_\lambda$                           & - \\
    DIAYN~\citep{Eysenbach:ICLR2019:DIAYN}       & Discrete            & Categorical                  & $q_\lambda$                           & -                         \\
    DIAYN (continuous)                           & Continuous ($\R^d$) & $\mathcal{N}(\mu(s), \Sigma(s))$     & $\mu(\cdot)$                & -                         \\
    DISCERN~\citep{WardeFarley:ICLR2019:DISCERN} & $=\S$ (e.g. image)  & Non-parametric                       & $\text{Embedding}(\cdot)$   & HER                       \\
    VISR~\citep{Hansen:ICLR2020:VISR}            & Continuous ($\R^d$) & $\text{vMF}(\mu(s), \kappa)$         & $\mu(\cdot)$                & SF                        \\\midrule
    VGCRL                       & Any                 & Any                                  & Any                         & P-HER or SF           \\ 

    \bottomrule
\end{tabular} %

\vspace*{-5pt}
\caption{%
    \small
    A summary of algorithms, all are optimized with the single objective in Eq.~\ref{eq:joint_op}.
    vMF stands for von Mise-Fisher distribution.
    DIAYN \citep{Eysenbach:ICLR2019:DIAYN},
    DISCERN \citep{WardeFarley:ICLR2019:DISCERN},
    VISR \citep{Hansen:ICLR2020:VISR} are special cases.
    For InfoGAIL~\citep{Li:2017:InfoGAIL}$^*$, we focus on the MI regularization objective $L_I(\pi, Q)$ only.
    Since they are under the same objective, learning techniques for goal-conditioned policy $\pi_z$
    such as successor features (SF)~\citep{Barreto:NIPS2017:SuccessorFeatures} and hindsight experience replay (HER)~\citep{Andrychowicz:NIPS2017:HER}
    can be adapted for more general settings within the VGCRL objective.
}
\label{tab:summary-algorithmsv2}
\vspace*{-5pt}
\end{table*}

If $z$ is continuous, a natural choice for $q_\lambda$ is a Gaussian, i.e. $\mathcal{N}(\mu(s), \Sigma(s))$,
where both $\mu$ and $\Sigma$ may be parameterized using any function approximators with a range of expressivities, from identity functions to deep neural networks.
Throughout the rest of the paper, we show how various simple choices for $q_\lambda$ lead to algorithms with different properties.

\paragraph{Goal-Conditioned RL as a Coarse Variational Approximation.}
\label{sec:identity_gauss}

A simple observation is that if we choose a fixed variational distribution,
such as $\mathcal{N}(s, \sigma^2 I)$ with $\sigma$ as a fixed hyperparameter and the goal space identical to the observation space ($\mathcal{Z} = \mathcal{S}$),
the RL objective in Eq.~\ref{eq:pi_op} becomes (see \Cref{sec:full-derivation} for mathematical details):
\begin{align}
    \mathcal F(\pi) =     \mathbb{E}_{z,s\sim \pi_\theta}\left[ -\textstyle\frac{1}{\sigma^2}\left\| z-s \right\|^2\right] + \mathrm{constant} \label{eq:goal_vge_op}.
\end{align}
It is straightforward to see that this recovers the objective of GCRL in Eq.~\ref{eq:goal_op} exactly (up to a constant),
where the distance function uses a squared loss. This provides a novel interpretation for GCRL algorithms as a \textit{variational empowerment algorithm with a hard-coded and fixed variational distribution}.
Given that no $q_\lambda$ parameters are adapted, this generally provides a very loose bound on MI;
however, prior work on GCRL shows that this RL objective, unlike empowerment-based, learns useful goal-reaching skills stably~\citep{kaelbling1993learning,Andrychowicz:NIPS2017:HER,pong2018temporal} thanks to a stationary reward function.
This suggests that GCRL and prior variational empowerment methods represent two ends of a spectrum, corresponding to the expressivity of the variational distribution used to approximately maximize mutual information, and neither of the two is perfect, with their own pros and cons.
Varying expressivity --- through the choices of $\mathcal{Z}$ and $q_\lambda$ --- and evaluating the qualities of learned goal spaces is a central theme of the next section.

\section{Goal-Conditioned RL as Variational Empowerment}
\vspace*{-3pt}

In this section, we discuss GCRL with representation learning, through the lens of variational empowerment: how the representation capacity leads to algorithms with different properties. We first derive two ``lost relatives'' of GCRL that only add minimal representation capacities but still result in interesting learning behaviors while keeping the stability of GCRL, and then discuss how we can study representation capacity in more general settings through varying smoothness constraints.
\subsection{Adaptive Variances for Relevance Determination}
\label{eq:ada_gcrl}
\vspace*{-5pt}

Given the observation in Section~\ref{sec:identity_gauss}, a straightforward modification to GCRL is to allow the variances to be learned, while keeping $\mu(s)=s$.
If we assume a global learned covariance, i.e. $q_\lambda(z|s)=\mathcal{N}(s, \Sigma)$, $\lambda=\{\Sigma\}$,
Eq.~\ref{eq:joint_op} gives us a novel variant of GCRL, which we call \emph{adaptive GCRL} (aGCRL).
The intuition behind this algorithm is the following: let us assume a simple diagonal covariance matrix;
during learning, this algorithm will quickly shrink $\sigma$ for the goal dimensions that the agent can reliably reach,
and will expand variances for the dimensions that the agent has a hard time to;
it therefore can identify and prioritize goal-reaching in feasible directions, discounting unfeasible ones, resembling properties of automatic relevance determination (ARD)~\citep{wipf2008new}.
\paragraph{Experiment: Automatic Controllability %
Determination on Windy PointMass.}
\begin{figure*}[t]
    \centering
    \begin{subfigure}[b]{0.20\linewidth} %
        \includegraphics[width=\linewidth,trim=20 20 20 20,clip]{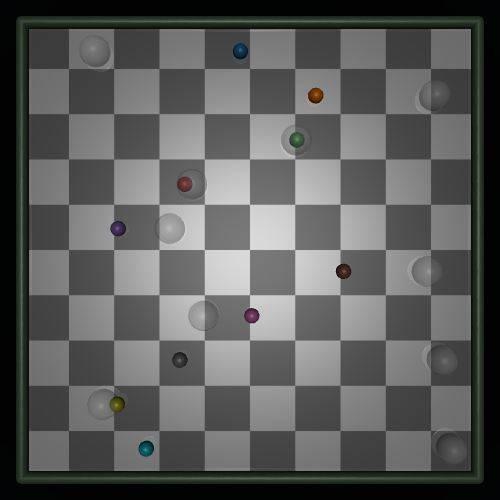}
        \caption{\small Windy PointMass.}
        \label{fig:env-pointmass}
    \end{subfigure}
    \begin{subfigure}[b]{0.255\linewidth} %
        \includegraphics[width=\linewidth]{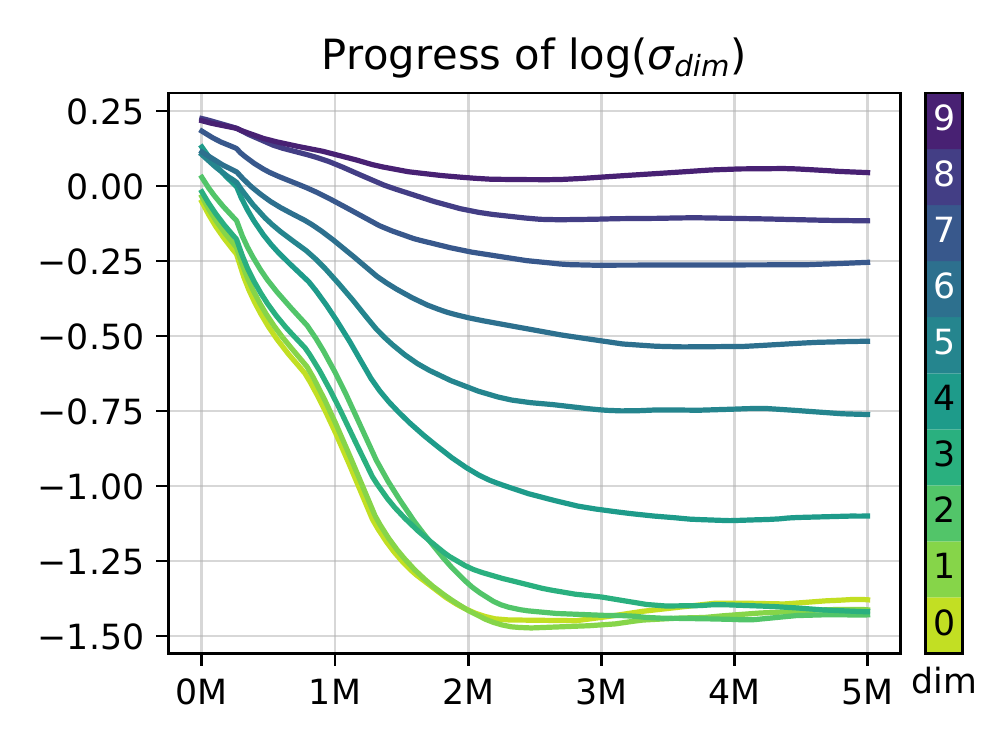}
        \caption{\small Learned variances (10D).}
        \label{fig:plot-adaptive-variance-b}
    \end{subfigure}
    \hfill
    \begin{subfigure}[b]{0.51\linewidth} %
        \includegraphics[width=\linewidth]{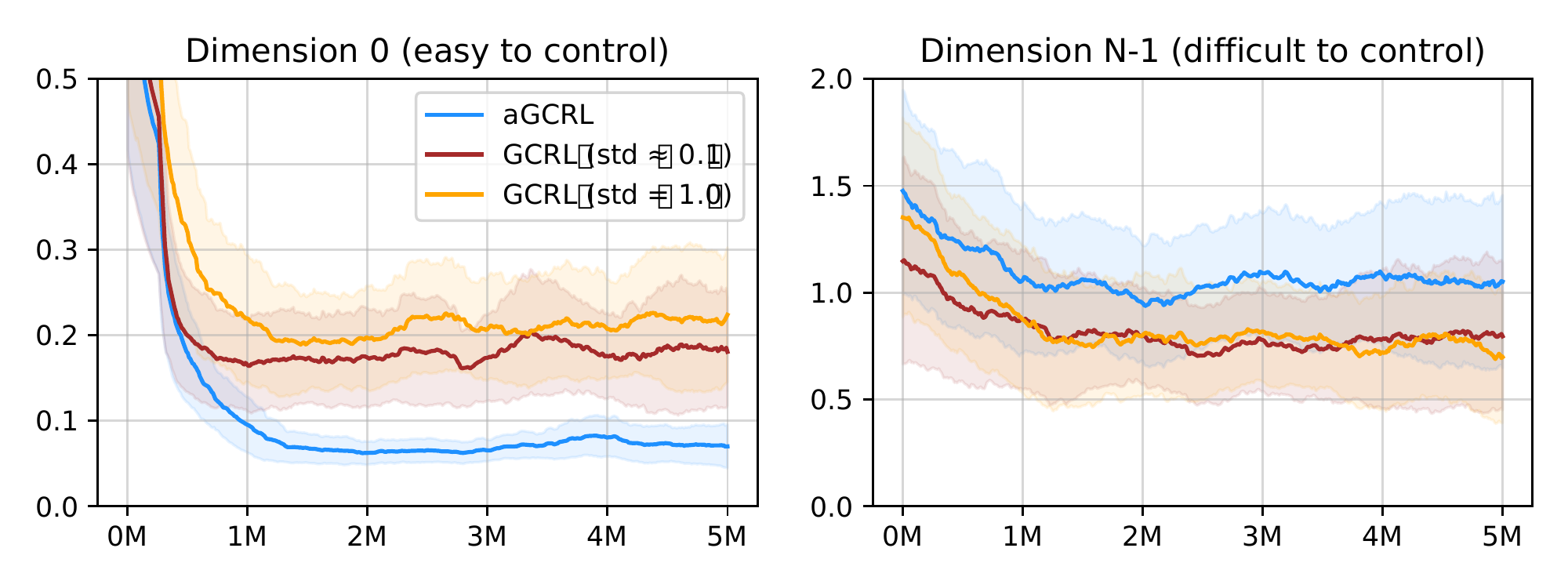}%
        \vspace*{-5pt}
        \caption{\small Goal reaching performance in the controllable dimension (dim 0)
        and the uncontrollable dimension (dim 9).
        The y-axis denotes the mean squared error between the goal location and achieve location.
        }
        \label{fig:plot-adaptive-variance-c}
    \end{subfigure}
    \vspace*{-5pt}
    \caption{\small
        Adaptive-variance GCRL. %
        The learned variance is clearly smaller for the easier (noiseless) dimension, %
        which makes goal reaching in controllable dimensions more focused than in uncontrollable ones. %
    }
    \label{fig:plot-adaptive-variance}
    \vspace*{-8pt}
\end{figure*}
We design a simple \textit{Windy PointMass} environment to study adaptive behaviors,
which is simulated in Mujoco~\citep{todorov2012mujoco}.
We assume a point mass in $N$-dimensional space (\Cref{fig:env-pointmass}), where some dimensions have random force perturbations
and therefore are difficult to control.
Such perturbations are often studied in the risk-sensitive RL literature~\citep{fox2015taming,maddison2017particle};
however, in our experiments, they serve to create different levels of controllability.
Our goal is to have GCRL automatically ignore dimensions that are not controllable and prioritize dimensions that are easy to control.
More details can be found in Appendix~\ref{sec:appendix_details}.

We evaluated goal-conditioned RL with an adaptive global diagonal variance term in Figure~\ref{fig:plot-adaptive-variance}.
Our results show that this simple modification to goal-based RL can accurately identify controllable dimensions in the state space.
For example,
    in a 2-dimensional windy pointmass environment,
    aGCRL recovered a smaller variance for the first dimension, $\sigma_x = 0.368$,
    and a larger variance for the second dimension, $\sigma_y = 1.648$,
which corresponds to having a reward function $r(s, z) = \|x - g_x\| / 0.368 + \|y - g_y\| / 1.648$ %
where $(x, y)$ is the position of the point mass and $z = (g_x, g_y)$ is the goal location.
A benefit of such adpative variance is that we can prioritize goal reaching in controllable dimensions;
in \Cref{fig:plot-adaptive-variance-c},
we can observe that aGCRL can reach goals in the controllable dimension (\eg dim 0) more quickly than the standard constant-variance GCRL baseline
on the 10-dimensional Windy PointMass environment,
showing the effectiveness of such
\emph{automatically learned reward functions that can ignore nuisance dimensions}.
\subsection{Adaptive Mean with Varying Expressivity}
\label{sec:linear_gauss}
\vspace*{-5pt}
The aGCRL variant
adapts variances but fixes the mean $\mu(s)$ to be $s$.
By using more expressive parameterizations, such as neural networks~\citep{Eysenbach:ICLR2019:DIAYN},
the algorithm can theoretically optimize a tighter lower-bound to MI.
However, as it gains more expressivity, interpretability and learning stability might be reduced.
We study a linear case, i.e. $q_\lambda(z|s)=\mathcal{N}(As, \Sigma)$ where $\lambda=\{A\}$ along with identity and NN cases,
and carefully evaluate this design choice. %
\begin{figure}[!b]   %
    \vspace*{-5pt}
    \centering
    \def\LinGCRLFigureHeight{78px}     %
    \includegraphics[height=\LinGCRLFigureHeight,trim=10 0 2000 10,clip]{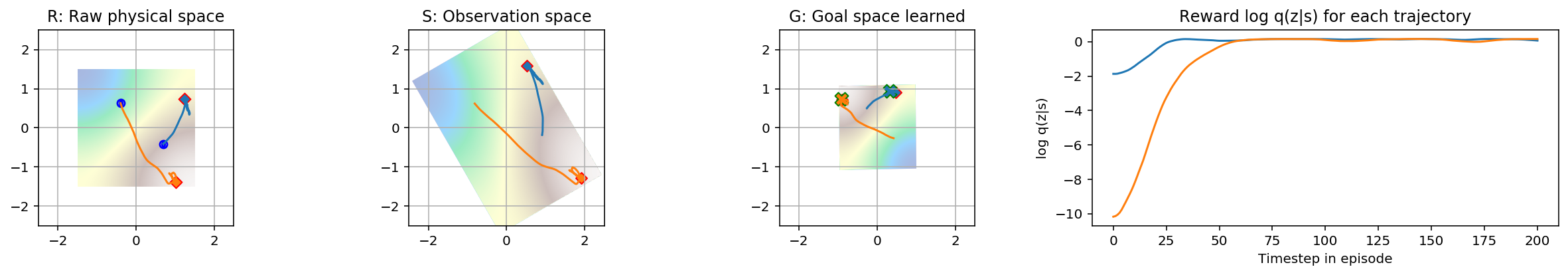}  %
    \includegraphics[height=\LinGCRLFigureHeight,trim=550 0 1450 10,clip]{./figures/randomproj-42.png}  %
    \includegraphics[height=\LinGCRLFigureHeight,trim=1100 0 850 10,clip]{./figures/randomproj-42.png}  %
        \\
    \vspace*{-15pt}
    \caption{ \small linGCRL on 2D point mass.
    Left: the underlying physics space.
    Middle: An agent's observation after a random linear projection.
    Right: A goal space recovered by $\mu = Ao$.
    The orange/green cross marks denote sampled goals $z \sim p(z)$,
    and blue dots are initial locations.
    }
    \label{fig:random-projection-1}
\end{figure}
\paragraph{Experiment: Recovering Intrinsic Dimensions of Variations with Linear GCRL.}
In this study, we design a simple 2D point mass with a random projection applied to the observation.
We use an affine transformation $W$ to generate the agent's observation $o = W s$ from a physics simulator's state $s$.
For example, when a raw state in 2D point mass (\Cref{fig:random-projection-1}) includes the $(x, y)$ location of the point mass,
the agent will instead receive an entangled, obfuscated observation: $o = (w_{11} x + w_{12} y, w_{21} x + w_{22} y, \ldots)$ which does not align with the action space.
We see whether a variant of VGCRL where $q(z | o) = \mathcal{N}(Ao, \Sigma)$, called \emph{linGCRL},
can recover the inverse of an underlying projection $A = W^{-1}$
when we use a rectangle-shaped 2D uniform prior $p(z)$. %
This resembles
PCA discovering principal components in and underlying true dimensionalities. %

Figure~\ref{fig:random-projection-1} shows an example where a unknown $2 \times 2$ random projection $W$ is applied.
The arena of the 2D point mass environment and two example trajectories are visualized:
the space of true state $s$ (left), observation $o = W s$ (middle),
and goal $z = (A \cdot W) s$.
We can see two intrinsic, orthogonal dimensions are recovered by the learned matrix $A$
such that the posteriors $z$ from marginal states match the prior distribution, but up to rotation and reflection.
This was also possible with more complex (e.g., $W \in \mathbb{R}^{10 \times 2}$) random projections.
We show that linGCRL can \emph{recover intrinsic dimensionalities of the state} from random projections.
\vspace*{-3pt}
\subsection{Spectral Normalization}             %
\label{sec:spectral-normalization}
\vspace*{-7pt}

\def\SpectralNormFigure{
\begin{figure}[t]
\centering
    \begin{subfigure}[b]{0.49\linewidth} %
        \includegraphics[width=\linewidth]{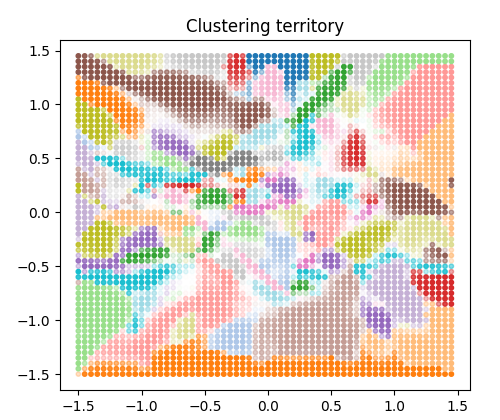}%
        \vspace*{-2pt}%
        \caption{\small Without SN.}
        \label{fig:clustering-without-sn}
    \end{subfigure}
    \begin{subfigure}[b]{0.49\linewidth} %
        \includegraphics[width=\linewidth]{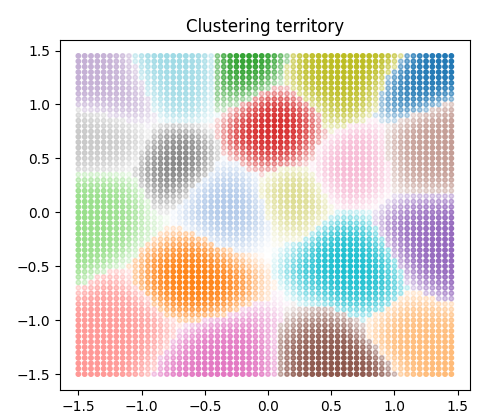}%
        \vspace*{-2pt}%
        \caption{\small With SN.}
        \label{fig:clustering-with-sn}
    \end{subfigure}
    \vspace*{-6pt}
  \caption{
      Visualization of the decision boundary of $q(z | s)$ on 2D point mass environment with $|\mathcal{G}| = 20$
      (See \textsection\ref{sec:spectral-normalization}).
  }
    \label{fig:spectral-norm-teaser}
    \vspace*{-9pt}
\end{figure}
}

If the variational posterior (``discriminator'') $q_\lambda (z | s)$ has high expressive power, for example when it is represented by a neural network, it can easily achieve maximum discriminability.
However, this can lead to a very suboptimal solution for the policy.
Because the marginal states as a result of latent goal and state pairs produced by the latent-conditioned policy are almost random and of poor quality in the early stage of training,
the discriminator might easily \emph{overfit} to such a near-random distribution of $(s, z)$, where $s \sim \pi_\theta(\cdot | \cdot, z)$.
This usually happens when in general $q_\lambda(z | s)$ is much easier to fit (especially given a high capacity of neural networks) than the policy $\pi_\theta$.
Figure~\ref{fig:spectral-norm-teaser}(a) shows a motivating example of $q_\lambda (z | s)$ learned
on a toy 2D point mass environment:
the landscape of $q(z | s)$ is highly non-smooth, %
which can hinder learning latent-conditioned skills due to a ill-posed reward structure.

\SpectralNormFigure

One way to alleviate this issue is to regularize the discriminator using Spectral Normalization~\citep{miyato2018spectral},
which has been very effective in stablizing Generative Adversarial Networks (GAN).
Intuitively speaking, spectral normalization enforces the discriminator to be a \emph{smooth} function by satisfying the Lipschitz continuity.
\paragraph{Experiment: Spectral Normalization.}             %

We study the behavior of variational empowerment algorithms on a toy 2D point mass environment,
for a purpose of simple analysis.
To simplify the analysis, we used $\mu(s) = (x, y)$, and a simple 2-layer MLP (with 128 hidden units).
\Cref{fig:spectral-norm-teaser} shows the decision boundary of the discriminator $q_\lambda(z | s)$, without and with Spectral Normalization (SN),
where the dimension of $z$ is 20 (or the number of skills).
We can observe that,
even though in both cases the empowerment objective $\mathcal F$
has converged to near-optimum value %
with a reasonably good discriminability
(\Cref{tbl:sn-discrete-compact}: top-1 accuracy of $q_\lambda(z | s)$ is $>90\%$),
the decision boundary is much more smooth with Spectral Normalization
and closer to interpretations of empowerment~\citep{Eysenbach:ICLR2019:DIAYN,mohamed2015variational}.
Without Spectral Normalization, the discriminator had to learn a highly non-smooth decision boundary
that is \emph{over-fit} to near-random data generated %
the premature policy $\pi_z$,
which would make joint optimization of $\pi_z$ and $q_\lambda(z | s)$ and discovery of meaningful
behaviors difficult.
Furthermore, spectral normalization can improve the performance
of variational empowerment algorithms in more challenging control tasks,
as will be discussed in \Cref{sec:experiments-her-lgr}.
These results confirm that \emph{high expressivity does not necessarily mean better performance},
and therefore that inductive biases or \emph{proper regularizations on the
posterior $q_\lambda(z | s)$
are important} for the performance of the algorithm and the quality of the goal representation learned.

\begin{table}[t]
    \centering
    \small
    \begin{tabular}{P{2.0cm} P{0.8cm}|lcccccc}
        \toprule
            {\small 2D Pointmass} & SN?        &  & $\mathcal F          $ & LGR($z$) \\
        \midrule %
            \multirow{2}{*}{$|G|=10$} & -          &  &          -0.38       &         0.94  \\
            {                       } & \Yes       &  &  \bf    {-0.14}      & \bf    {0.96} \\
        \midrule
            \multirow{2}{*}{$|G|=20$} & -          &  &          -0.86       &         0.91  \\
            {                       } & \Yes       &  &  \bf    {-0.24}      & \bf    {0.96} \\
        \midrule
            \multirow{2}{*}{$|G|=50$} & -          &  &  \bf    {-1.02}      & \bf    {0.83} \\
            {                       } & \Yes       &  &         {-1.15}      &         0.78  \\
        \bottomrule
    \end{tabular}
    \vspace*{-4pt}
    \captionof{table}{ %
        {DIAYN v.s.~DIAYN + Spectral Normalization (SN) on 2D PointMass (See \Cref{fig:spectral-norm-teaser}).
        Please refer to \Cref{sec:experiments-her-lgr} for details of the metrics and discussions.
        }
    }
    \label{tbl:sn-discrete-compact}
    \vspace*{-6pt}
\end{table}

\vspace*{-2pt}
\section{Variational Empowerment as Goal-Conditioned RL}
\vspace*{-5pt}
\begin{figure*}[t]
    \captionsetup[subfigure]{aboveskip=-1pt,belowskip=2pt}
    \begin{subfigure}[t]{0.49\linewidth} %
        \includegraphics[width=\linewidth]{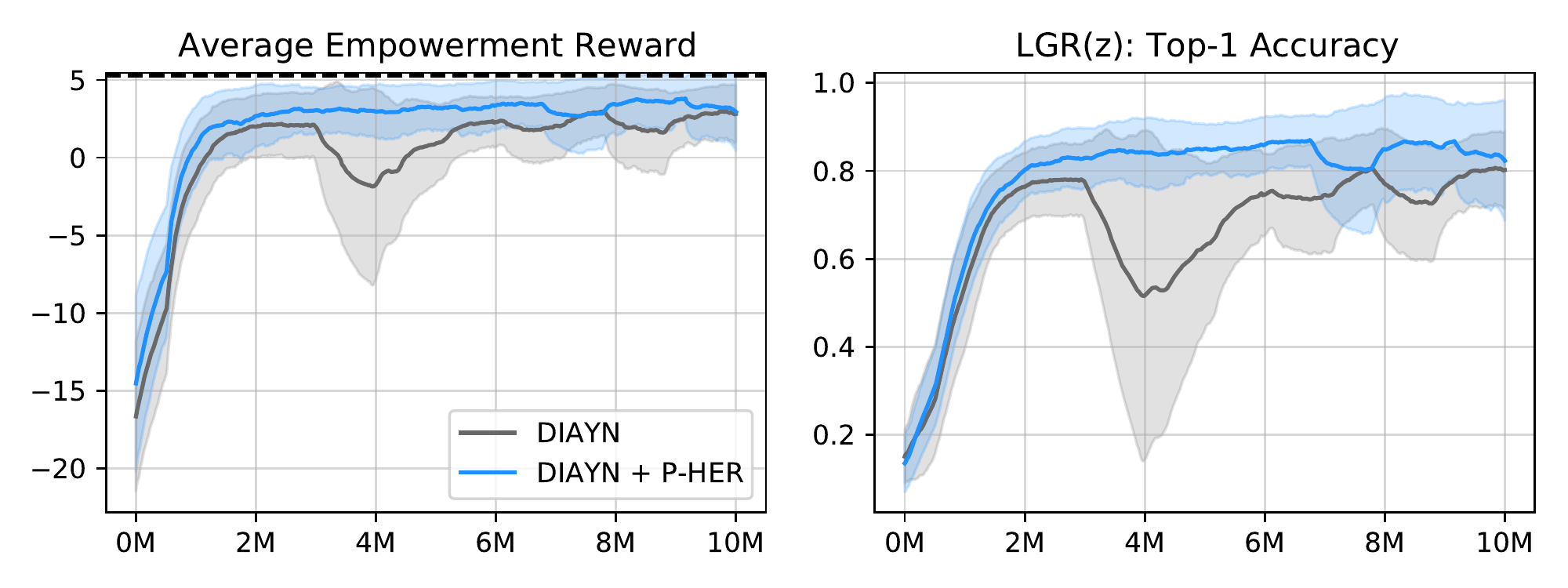}
        \caption{ HalfCheetah-v3 (Discrete, $|\mathcal G| = 200$)}
    \end{subfigure}
    \begin{subfigure}[t]{0.49\linewidth} %
        \includegraphics[width=\linewidth]{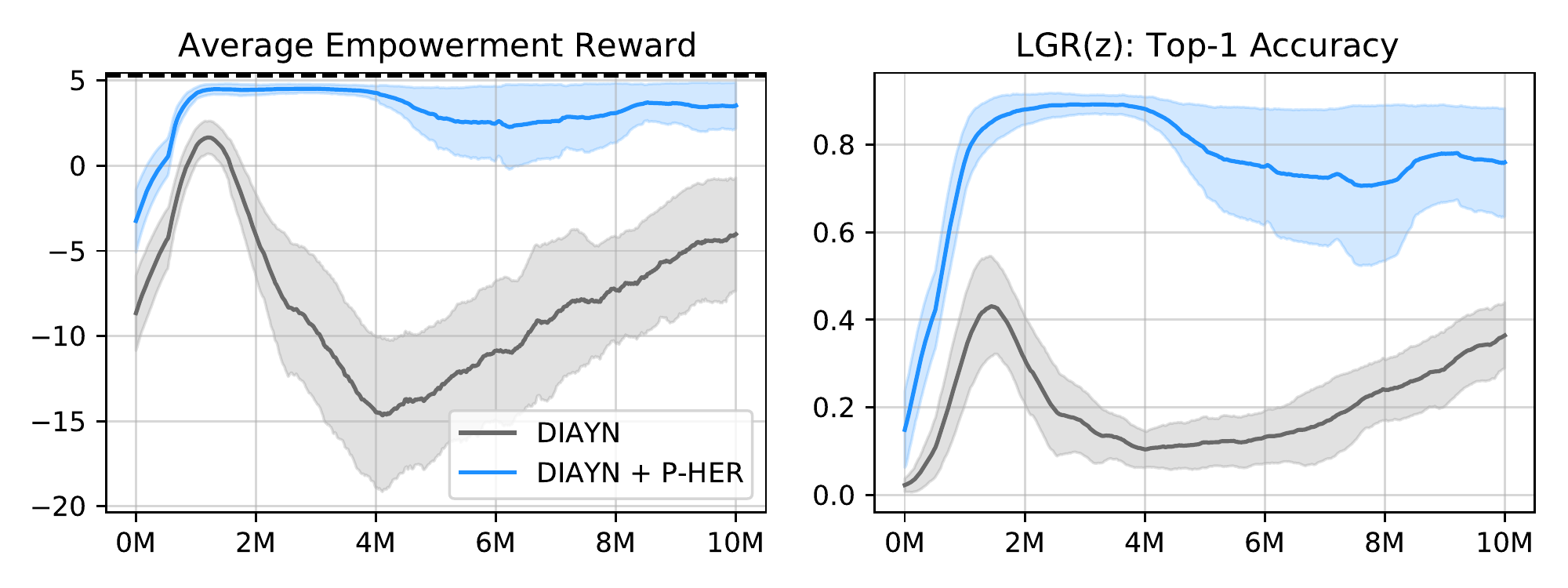}
        \caption{ Humanoid-v3 (Discrete, $|\mathcal G| = 200$)}
    \end{subfigure}
    \\[-2pt]
    \begin{subfigure}[t]{0.49\linewidth} %
        \includegraphics[width=\linewidth]{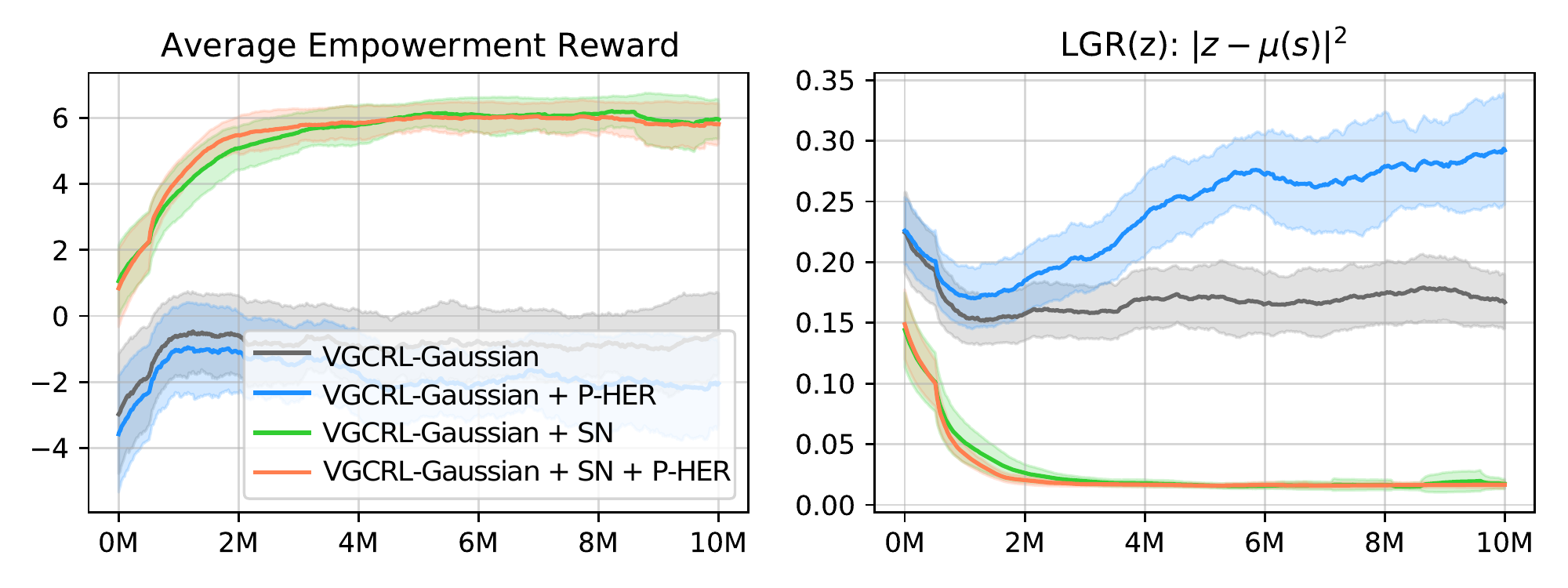}
        \caption{ HalfCheetah-v3 (Gaussian, $|\mathcal G| = 5$)}
    \end{subfigure}
    \begin{subfigure}[t]{0.49\linewidth} %
        \includegraphics[width=\linewidth]{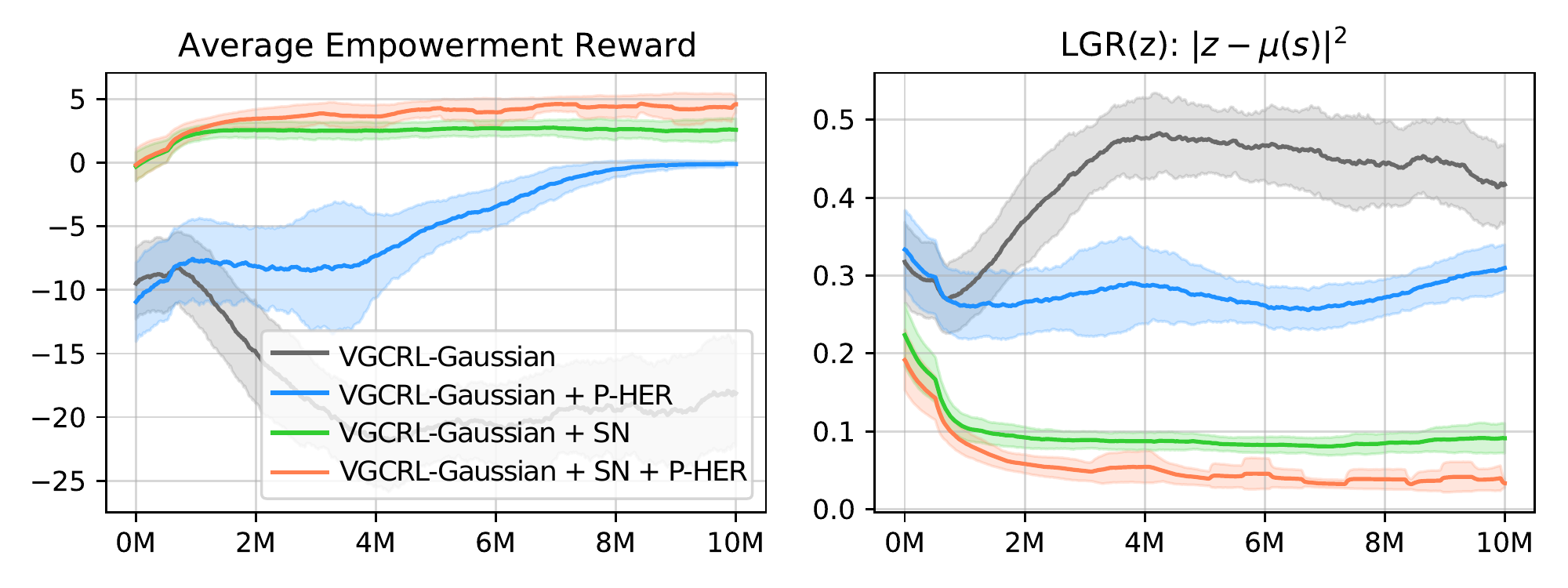}
        \caption{ Ant-v3 (Gaussian, $|\mathcal G| = 5$)}
    \end{subfigure}
    \vspace*{-7pt}
    \caption{  %
        Learning curves (selected) of variational empowerment methods with
        \textit{discrete} ((a)-(b), see \Cref{tbl:result-pher}) and
        \textit{continuous} ((c)-(d), see \Cref{tbl:summary-key-results-compact})
        goal spaces.
        We can see that Posterior HER (P-HER) makes optimization of the MI objective faster and more stable,
        especially when the goal space is large and/or goal-conditioned policy is difficult to learn.
        The use of spectral normalization (SN, \Cref{sec:spectral-normalization})
        provides performance gains for VGCRL with Gaussian discriminators.
        The higher LGR$(z)$ is, the better in discrete cases (accuracy; (a)-(b));
        the lower LGR$(z)$ is, the better in continuous cases (distance; (c)-(d)).
        Learning curves are averaged over 3 random seeds.
        More plots can be found in \Cref{sec:appendix_more_results}.
    }\label{fig:plot-pher}
\end{figure*}

We have seen that goal-conditioned RL can be viewed as a special case of the unified optimization objective in \Cref{eq:joint_op}.
A natural question that follows is whether training methods that work well for GCRL can also be used within the more general VGCRL framework.
We answer in the affirmative, deriving an extension to a goal-relabeling technique (HER) and an evaluation metric for variational empowerment algorithms.
We present some techniques that are helpful for learning VGCRL.   %
\subsection{P-HER: Posterior Hindsight Experience Replay}
\label{sec:pher}
\vspace*{-5pt}

As discussed in Section~\ref{sec:bg_gcrl}, Hindsight Experience Replay (HER)~\citep{Andrychowicz:NIPS2017:HER}
substantially improves off-policy learning of goal-conditioned policies and value functions.
Inspired by the mathematical connection to GCRL, %
we can derive an equivalent of HER or goal relabeling in the context of VGCRL.
Specifically,
we relabel $z$ for fitting the policy $\pi_\theta(a | s, z)$ with the relabeled goal $\mathbb{S}(s_{0:T})$ derived from the final state $s_T$:
\vspace*{-4pt}
\begin{align}
    \mathbb{S}(s_{0:T}) \sim q_\lambda(z \mid s_T).
\end{align}
We call this relabeling technique \emph{Posterior HER} (\textbf{P-HER}), as it can be seen as an application of posterior sampling~\citep{hausman2018learning,rakelly2019efficient} to HER.
Since $q_\lambda$ is non-stationary, we always use the \emph{up-to-date} estimate of $q_\lambda(z | s_T)$ with the latest parameter $\lambda$ of posterior $q_\lambda$.
This is the biggest difference to the vanilla HER where the goal mapping function is fixed.
We note that other state sampling distributions of HER, e.g., uniform over $s_{0:T}$ or $s_{t>k}$ can also be trivially supported.
For fitting the discriminator $q_\lambda (z | s)$, we do not relabel the goal.
Similarly to HER in GCRL, this technique can be viewed as a curriculum for enabling parameterized policy optimization to focus on high-reward regions.
This technique can \emph{accelerate optimization steps} for $\pi_\theta$ (Equation~\ref{eq:pi_op})
in the same spirit of HER
when latent goal reaching is non-trivial, especially in high-dimensional state spaces and with immaturely-shaped
reward functions.

\subsection{Latent Goal Reaching: A Metric for MI-Based Empowerment Algorithms}
\label{sec:lgr}
\vspace*{-5pt}

One limitation in mutual information-based RL~\citep{Eysenbach:ICLR2019:DIAYN,Sharma:ICLR2020:DADS} is lack of objective evaluation metrics.
Previous works directly evaluate qualitative results of learned behaviors,
but quantitative metrics are limited to discriminator rewards (\ie, $\mathbb E[ \log q(z | s) - p(z) ]$)
which can saturate and do not always correspond to the quality of learned behaviors,
or downstream task evaluations such as exploration bonus and pre-trained primitives for hierarchical RL~\citep{%
florensa2017stochastic,Eysenbach:ICLR2019:DIAYN,Sharma:ICLR2020:DADS,Hansen:ICLR2020:VISR}.

By contrast, goal-conditioned RL has a clear metric: define a sample of goals and evaluate average goal-reaching performance.
Since we demonstrated how MI-based RL is closely related to GCRL,
we propose Latent Goal Reaching (LGR) as a new metric for evaluating MI-based algorithms such as DIAYN or DADS.
The procedure is described in Algorithm~\ref{algorithm:lgr}.
\emph{It allows us to evaluate MI-based RL as just another goal-reaching problem}:
we measure how accurately the goal-conditioned policy can reach the given goal states of interest.
This metric measures the quality of both the discriminator (\ie, goal representation) and the goal-conditioned policy.
We note the range of this metric is not dependent on the choice of distribution family of $q(\cdot)$,
allowing comparison across different types of $q(z | s)$.

\begin{algorithm}[H]   %
    \small
    \SetAlgoLined
    \textbf{Input}: Target states $s^{1:N}$, trained $\pi_\theta(a|s,z),q_\lambda(z|s)$\\
    \textbf{Output}: Average distance $\bar{d}$ to goal states %
    \\
     \For{$i\gets1$ \KwTo $N$}{
         Embed target state into a goal: $z^i\gets \mathbb{E}\left[q_\lambda(\cdot|s^i)\right]$ \\
         Run $\pi(\cdot|\cdot,z^i)$ for $T$ time steps, observe final state $s^i_T$ \\
         Compute $d(s^i, s^i_T)$ ~~(\eg, squared distance)
        }
        Report the average over $N$ episodes: $\bar{d} = \sum_i d(s^i, s^i_T) / N$ %
    \caption{Latent Goal Reaching Metric: LGR($s$)}
    \label{algorithm:lgr}
\end{algorithm}
\vspace*{-10pt}

The value of LGR($s$) metric depends on the choice of target states.
In locomotion control tasks~\citep{brockman2016gym},
we would often want to discover and learn behaviors where the robot can walk or move~\citep{Eysenbach:ICLR2019:DIAYN,Sharma:ICLR2020:DADS}.
To evaluate this, one can generate diverse target states
of moving in different directions and at different velocities from expert policies.
In such cases we can compute the distance $d(s^i, s_T^i)$ between target and achieved states
with respect to velocity dimensions (\eg, velocity in $x$ and $y$ axis).
We call this variant of the LGR metric as LGR$_v(s)$.
Details of target state generation used in our experiments are given in \Cref{sec:appendix_details}.

We also consider the LGR($z$) metric which measures the goal reaching performance in the latent goal space.
For discrete $z$, this simply can be the top-1 accuracy of the discriminator $q(z | s)$ with respect to the true goal $z \sim p(z)$.
For continuous $z$,
LGR($z$) is the squared distance $\|z - \argmax q(z|s) \|^2$ between the goal and the mode of the discriminator $\argmax q(z | s)$.
We note that this metric agrees the state distance metric used in Laplacian embedding~\citep{wu2018laplacian}, 
i.e., $\|\phi(s) - \phi(h^{-1}(z))\|$,
with a state embedding $\phi(s): = \argmax_z q(z|s)$,
where $h^{-1}(z)$ is defined to be an arbitrary state $s$ that is associated with latent $z$,
which is in our case the marginal state from the latent-conditioned policy $\pi_\theta$.
A difference is that \cite{wu2018laplacian} use contrastive learning whereas
VGCRL maximizes likelihood to learn the representation $q$.

\subsection{Experiments: Posterior HER and Latent Goal Reaching}
\label{sec:experiments-her-lgr}
\vspace*{-5pt}
\begin{table*}[!h]
\footnotesize
\setlength{\tabcolsep}{4pt}
\centering
\begin{tabular}{ll|cHcc|cHcc|cHcc}
\toprule
                    & Method        & \multicolumn{4}{c}{HalfCheetah}                 & \multicolumn{4}{c}{Ant}                          &  \multicolumn{4}{c}{Humanoid}
\\
$|\G|$              &               & $\mathcal F$   &       LGR($s$) &  LGR$_v$($s$)  &       LGR($z$) & $\mathcal F$   &       LGR($s$) &  LGR$_v$($s$)  & LGR($z$)       & $\mathcal F$   &       LGR($s$) &   LGR$_v$($s$) &        LGR($z$)   \\
\midrule
10                  & DIAYN         & \bf      1.608 & \bf    208.289 & \bf      0.800 & \bf      0.963 &         -0.835 & \bf    165.028 &          0.529 &          0.806 &          1.261 & \bf    38308.1 &          0.523 &          0.922    \\ 
                    & DIAYN + P-HER &          1.372 &        293.950 &          1.424 &          0.934 & \bf     -0.049 &        176.905 & \bf      0.486 & \bf      0.889 & \bf      1.856 &        40193.4 & \bf      0.312 & \bf      0.953    \\ 
\midrule                              
20                  & DIAYN         &          1.732 & \bf    330.125 & \bf      1.125 & \bf      0.920 &         -1.308 &        168.874 &          0.610 &          0.763 &          0.713 & \bf    36398.1 &          0.315 &          0.768    \\ 
                    & DIAYN + P-HER & \bf      1.852 &        411.069 &          1.214 &          0.891 & \bf     -1.288 & \bf    155.585 & \bf      0.515 & \bf      0.823 & \bf      2.251 &        41891.6 & \bf      0.183 & \bf      0.922    \\
\midrule                              
50                  & DIAYN         &          1.475 & \bf    225.862 & \bf      0.673 &          0.827 &         -3.812 &        170.345 & \bf      0.402 &          0.523 &         -1.158 & \bf    38632.8 & \bf      0.268 &          0.549    \\ 
                    & DIAYN + P-HER & \bf      1.699 &        232.120 &          0.704 & \bf      0.834 & \bf     -2.171 & \bf    166.517 &          0.637 & \bf      0.750 & \bf      2.848 &        43504.1 &          0.545 & \bf      0.891    \\
\midrule                              
200                 & DIAYN         &          2.854 & \bf    231.073 & \bf      0.698 &          0.801 &         -8.450 & \bf    136.580 & \bf      0.396 &          0.113 &         -4.396 & \bf    33174.9 & \bf      0.208 &          0.337    \\ 
                    & DIAYN + P-HER & \bf      3.357 &        273.566 &          0.814 & \bf      0.844 & \bf     -7.047 &        184.991 &          0.555 & \bf      0.263 & \bf      3.448 &        33838.6 &          0.866 & \bf      0.766    \\
\midrule                              
1000                & DIAYN         &         -8.286 &        315.127 &          1.156 &          0.176 &        -16.795 &        135.171 &          0.434 &          0.005 &         -8.914 & \bf    36686.1 & \bf      0.325 &          0.101    \\ 
                    & DIAYN + P-HER & \bf     -4.424 & \bf    218.390 & \bf      0.510 & \bf      0.361 & \bf    -10.941 & \bf    134.100 & \bf      0.322 & \bf      0.028 & \bf      3.395 &        41381.4 &          1.569 & \bf      0.762    \\
    \bottomrule
\end{tabular}
\vspace*{-5pt}
\caption{Evaluation of Latent Goal-Reaching Metric on MuJoCo control suites, after a total of 10M environment steps of training.
    $\mathcal F$ is the (average) empowerment reward, $\mathcal{F} = \mathbb{E}_z [\log q_\lambda(z | s) - p(z)]$.
LGR$_v$($s$) is the squared error in observation space (the lower, the better) with respect to velocity dimensions between the marginal state and the target state,
and LGR($z$) denotes the accuracy of top-1 classification of the discriminator $q_\lambda(z|s)$ (the higher, the better, max 1.0).
}
\label{tbl:result-pher}
\vspace*{-3pt}
\end{table*}
\vspace*{1pt}
\def\tablefullflag{0}
\providecommand{\TwoRows}[1]{\multirow{2}{*}{#1}}
\providecommand{\FourRows}[1]{\multirow{4}{*}{#1}}
\providecommand{\Gvertical}[1]{\multirow{8}{*}{ \rotatebox{90}{$|\G| = #1$} }}

\begin{table*}[t]
    \linespread{0.95}
    \hspace*{-17pt}
    \footnotesize
\centering
\setlength\tabcolsep{2.5pt}
\begin{tabular}{cccc|p{0pt}cHcc|cHcc|cHcc} %
    \toprule
                    & \TwoRows{$q_\lambda (z|s)$                     } & \hspace*{-10pt}\TwoRows{P-HER?}     & \TwoRows{SN?}& & \multicolumn{4}{c}{HalfCheetah}                   & \multicolumn{4}{c}{Ant}                          &  \multicolumn{4}{c}{Humanoid}      \\
                ~   & {                                                                  } & {             } &              & & $\mathcal F$    &  LGR($s$)      & LGR$_v$($s$)   & LGR($z$)       & $\mathcal F$   &  LGR($s$)      & LGR$_v$($s$)   & LGR($z$)       &  $\mathcal F$  & LGR($s$)       & LGR$_v$($s$)   & LGR($z$)       \\ \midrule
\ifnum0\tablefullflag>0 %
    \Gvertical{2}   & \TwoRows{$\mathcal{N}(\mu(s), \text{fixed}^2)$                     } & -            {} & -           &  &         0.305 &        296.399 &          0.900 &          0.166 &         -0.128 &        150.310 &          0.398 &          0.242 &         -0.047 &        34793.4 &          2.394 &          0.199    \\ 
                    & {                                                                  } & \Yes         {} & -           &  &         0.339 &        287.019 &          0.837 &          0.139 &         -0.110 &        152.557 &          0.678 &          0.306 &         -0.082 &        35617.5 &          1.505 &          0.194    \\ 
    \cmidrule{2-17} & \FourRows{$\mathcal{N}(\mu(s), \Sigma(s)^2)$                       } & -            {} & -           &  &         0.830 & \bf    228.036 &          1.177 &          0.063 &         -4.669 &        132.998 &          0.478 &          0.265 &          0.677 & \bf    31160.8 &          0.393 &          0.080    \\ 
                    & {                                                                  } & \Yes         {} & -           &  &         0.403 &        272.550 &          0.720 &          0.079 &         -4.575 &        151.936 & \bf      0.289 &          0.263 &          2.019 &        31586.3 &          0.910 &          0.027    \\ 
                    & {                                                                  } & -            {} & \Yes        &  &         2.653 &        314.283 &          1.017 &          0.011 &          2.060 &        138.529 &          0.453 &          0.038 &          2.511 &        35293.8 &          0.225 &          0.012    \\ 
                    & {                                                                  } & \Yes         {} & \Yes        &  &\bf      2.724 &        324.174 &          1.074 & \bf      0.009 & \bf      2.352 & \bf    126.262 &          0.511 & \bf      0.023 & \bf      2.549 &        37243.4 & \bf      0.199 & \bf      0.012    \\ 
    \cmidrule{2-17} & \TwoRows{GMM ($K=8$)                                               } & -            {} & -           &  &         0.883 &        245.145 & \bf      0.707 &          0.188 &         -4.344 &        168.075 &          0.640 &          0.360 &          1.141 &        38213.0 &          1.637 &          0.072    \\ 
                    & {                                                                  } & \Yes         {} & -           &  &         1.183 &        484.277 &          2.032 &          0.181 &         -3.436 &        165.852 &          0.432 &          0.356 &          2.076 &        31623.4 &          0.993 &          0.026    \\ 
    \midrule
\fi
\ifnum0\tablefullflag>-1 %
    \Gvertical{5}   & \TwoRows{$\mathcal{N}(\mu(s), \text{fixed}^2)$                     } & -            {} & -           &  &         0.932 &        228.504 &          1.005 &          0.159 &         -0.590 &        184.930 &          1.005 &          0.382 &          0.239 &        49001.5 &          1.461 &          0.202    \\ 
                    & {                                                                  } & \Yes         {} & -           &  &        -0.142 &        259.051 &          1.273 &          0.360 &          0.140 &        196.357 &          2.449 &          0.300 &          0.020 &        36964.5 &          1.452 &          0.244    \\ 
    \cmidrule{2-17} \fi & \FourRows{$\mathcal{N}(\mu(s), \Sigma(s)^2)$                   } & -            {} & -           &  &        -0.731 &        378.998 &          1.251 &          0.172 &        -18.490 & \bf    115.022 & \bf      0.306 &          0.427 &         -3.597 &        37426.1 &          0.538 &          0.147    \\ 
                    & {                                                                  } & \Yes         {} & -           &  &        -2.161 &        262.668 &          1.132 &          0.289 &         -0.108 &        258.082 &          2.423 &          0.303 &          1.207 &        44776.0 &          0.206 &          0.074    \\ 
                    & {                                                                  } & -            {} & \Yes        &  &\bf      5.856 &        222.438 & \bf      0.604 &          0.019 &          2.548 &        206.547 &          0.925 &          0.091 &          4.509 &        43146.3 &          0.460 &          0.040    \\ 
                    & {                                                                  } & \Yes         {} & \Yes        &  &         5.803 &        328.152 &          1.352 & \bf      0.017 & \bf      4.349 &        164.416 &          0.463 & \bf      0.039 & \bf      5.203 &        38002.7 & \bf      0.203 & \bf      0.026    \\ 
\ifnum0\tablefullflag>-1 \cmidrule{2-17} & \TwoRows{GMM ($K=8$)                           } & -            {} & -           &  &        -2.646 & \bf    221.274 &          0.766 &          0.325 &        -16.196 &        134.348 &          0.367 &          0.486 &         -3.576 &        38033.3 &          0.231 &          0.198    \\ 
                    & {                                                                  } & \Yes         {} & -           &  &        -3.091 &        278.247 &          1.065 &          0.404 &         -2.874 &        201.952 &          2.794 &          0.325 &          3.526 & \bf    34782.6 &          0.581 &          0.043    \\ 
\fi                                                                                                                            
\ifnum0\tablefullflag>0 %
    \midrule                                                                                                                   
    \Gvertical{10}                      & \TwoRows{$\mathcal{N}(\mu(s), \text{fixed}^2)$ } & -            {} & -           &  &        -0.145 &        222.187 &          0.855 &          0.346 &         -1.719 &        131.568 &          0.674 &          0.508 &          0.246 &        50502.0 &          0.704 &          0.237    \\ 
                                        & {                                              } & \Yes         {} & -           &  &        -0.825 &        344.320 &          0.866 &          0.407 &         -0.276 &        293.498 &          2.309 &          0.330 &          0.125 & \bf    30696.7 &          0.708 &          0.239    \\ 
    \cmidrule{2-17}                     & \FourRows{$\mathcal{N}(\mu(s), \Sigma(s)^2)$   } & -            {} & -           &  &        -3.688 &        384.922 &          1.381 &          0.384 &         -6.709 &        166.072 &          0.745 &          0.425 &         -3.399 &        40792.9 &          0.313 &          0.221    \\ 
                                        & {                                              } & \Yes         {} & -           &  &        -3.582 & \bf    219.542 & \bf      0.640 &          0.388 &         -0.190 &        282.134 &          3.989 &          0.324 &         -3.618 &        43749.2 & \bf      0.244 &          0.111    \\ 
                                        & {                                              } & -            {} & \Yes        &  &         3.840 &        324.391 &          1.175 &          0.180 &          0.721 &        205.621 &          0.974 &          0.240 & \bf      8.134 &        36740.2 &          1.275 & \bf      0.061    \\ 
                                        & {                                              } & \Yes         {} & \Yes        &  &\bf      4.975 &        328.543 &          0.874 & \bf      0.162 & \bf      2.467 &        151.445 &          0.674 & \bf      0.184 &          6.349 &        37585.3 &          0.262 &          0.072    \\ 
                        \cmidrule{2-17} & \TwoRows{GMM ($K=8$)                          } & -            {} & -           &  &        -5.137 &        356.233 &          1.250 &          0.404 &        -25.121 & \bf    126.206 & \bf      0.307 &          0.534 &          1.885 &        39255.1 &          1.543 &          0.238    \\ 
                                        & {                                              } & \Yes         {} & -           &  &        -6.162 &        284.191 &          0.835 &          0.399 &         -3.582 &        286.026 &          2.396 &          0.348 &          3.267 &        43158.8 &          0.422 &          0.082    \\ 
\fi
\ifnum0\tablefullflag>0 %
    \midrule
    \Gvertical{20}  & \TwoRows{$\mathcal{N}(\mu(s), \text{fixed}^2)$                     } & -            {} & -           &  &        -2.024 &        256.742 &          0.901 &          0.438 &         -3.206 &        133.231 & \bf      0.486 &          0.498 &         -0.656 &        41267.2 &          0.622 &          0.320    \\ 
                    & {                                                                  } & \Yes         {} & -           &  &        -1.848 &        277.402 &          0.953 &          0.422 &         -0.527 &        304.335 &          3.038 &          0.331 &         -0.295 &        38000.6 &          0.461 &          0.295    \\ 
    \cmidrule{2-17} & \FourRows{$\mathcal{N}(\mu(s), \Sigma(s)^2)$                       } & -            {} & -           &  &        -3.754 & \bf    129.309 & \bf      0.481 &          0.377 &         -2.662 &        151.189 &          0.958 &          0.353 &          1.705 & \bf    34799.1 &          2.115 &          0.274    \\ 
                    & {                                                                  } & \Yes         {} & -           &  &        -4.704 &        365.717 &          1.648 &          0.385 &         -2.813 &        301.216 &          1.000 &          0.352 &          2.149 &        56381.8 &          0.731 &          0.246    \\ 
                    & {                                                                  } & -            {} & \Yes        &  &        -0.176 &        280.557 &          1.054 &          0.318 &         -1.624 &        174.646 &          0.716 &          0.355 & \bf      7.176 &        39763.5 &          1.666 & \bf      0.191    \\ 
                    & {                                                                  } & \Yes         {} & \Yes        &  &\bf      0.727 &        262.139 &          1.066 & \bf      0.294 & \bf     -0.503 & \bf    117.467 &          0.496 & \bf      0.320 &         -0.350 &        42427.7 & \bf      0.460 &          0.340    \\ 
    \cmidrule{2-17} & \TwoRows{GMM ($K=8$)                                               } & -            {} & -           &  &        -6.294 &        238.938 &          1.254 &          0.394 &        -11.060 &        178.428 &          0.805 &          0.370 &          1.579 &        39450.0 &          2.280 &          0.298    \\ 
                    & {                                                                  } & \Yes         {} & -           &  &       -10.647 &        395.075 &          1.725 &          0.397 &        -13.392 &        265.111 &          1.340 &          0.377 &          2.339 &        39036.9 &          1.740 &          0.284    \\ 
\fi %
    \bottomrule
\end{tabular} %

\ifnum0\tablefullflag>0 %
    \caption{%
        An extended version of \Cref{tbl:summary-key-results-compact}.
        We present a VGCRL-Gaussian variant where the variance is not learned but kept constant 
        (fixed, \eg $\log \sigma$ = 0) and a variant where the variance is learned as a function of state $s$.
        VGCRL-GMM is when a Gaussian Mixture Model is used for the discriminator instead of a Gaussian distribution,
        where means, covariances, and mixture weights are learned through the neural network.
    }%
    \label{tbl:summary-key-results}
\else
    \vspace*{-5pt}
    \caption{%
        Comparison of continuous variants of VGCRL, where the dimension of the goal space is $|\mathcal G| = 5$.
        SN denotes Spectral Normliazation~(\Cref{sec:spectral-normalization}).
        LGR($z$) is the goal reaching performance in the latent space (the lower, the better).
        A full table containing more comprehensive comparison %
        and corresponding plots
        can be found in \Cref{sec:appendix_more_results} (\Cref{tbl:summary-key-results}).
    }%
    \label{tbl:summary-key-results-compact}
    \vspace*{-5pt}
\fi %
\end{table*}

In this section, we evaluate the performance of several variants of VGCRL on standard locomotion tasks~\citep{brockman2016gym}.
We consider both a discrete latent space, analogous to that used by DIAYN~\citep{Eysenbach:ICLR2019:DIAYN}, %
and a continuous latent space, %
where the variational posterior  $q_\lambda(z | s)$ chosen to be a
Gaussian posterior (VGCRL-Gaussian), with either learnable or fixed variances,
or Gaussian Mixtures (VGCRL-GMM).
To evaluate the performance, we report the following metrics:
(1) the empowerment objective $\mathcal{F} = \mathbb{E}_z[\log q(z | s) - p(z)]$,
(2) LGR($z$): the latent goal reaching metric,
and (3) LGR$_v$($s$): the latent goal reaching metric
with respect to velocity dimensions (\Cref{sec:lgr}).

We first observe that P-HER can accelerate and improve learning (\Cref{fig:plot-pher}, \Cref{tbl:result-pher,tbl:summary-key-results-compact}).
Such improvements are significant in high-dimensional goal spaces (\eg, $|\G| = 200$)
and more difficult control tasks such as Ant or Humanoid with high dimensionalities,
in which both the discriminator and the latent-conditioned policy are difficult to learn.
Because an optimization of goal-conditioned policy (Eq.~\ref{eq:pi_op}) is more difficult than discriminators (Eq.~\ref{eq:lambda_op}),
relabeling of goal can greatly accelerate RL, which also results in better discriminability.
As shown in \Cref{tbl:summary-key-results-compact,tbl:summary-key-results},
\textbf{P-HER} can improve
not only the optimization objective but also other metrics
such as discriminator's accuracy and goal reaching performance
across different design choices and environments.

\emph{Moreover, when Spectral Normalization (\Cref{sec:spectral-normalization}) is applied,
we observe a significant improvement in terms of the learning progress and evaluation metrics} (\Cref{fig:plot-pher}, \Cref{tbl:summary-key-results-compact}).
As shown in \Cref{fig:plot-pher},
while there are some progress with vanilla VGCRL-Gaussian (or VGCRL-GMM),
the optimization objective $\mathcal F$ as well as evaluation metrics do not improve as much as the one with SN,
despite the \emph{bigger expressivity} due to no constraint on $q_\lambda(z|s)$.
We can also see that with SN (and P-HER as well), the distance between achieved and desired goal is much lower.
More experimental results and discussions can be found in \Cref{sec:appendix_more_results}.

\vspace*{-3pt}
\section{Conclusion}
\label{sec:conclusion}
\vspace*{-5pt}

Our variational GCRL (VGCRL) framework unifies unsupervised skill learning methods based on variational empowerment~\citep{Eysenbach:ICLR2019:DIAYN,Gregor:ICLRWorkshop2017:VIC,Sharma:ICLR2020:DADS} with goal-conditioned RL (GCRL) methods, allowing us to transfer techniques and insights across both types of approaches.
Viewing GCRL as variational empowerment, we derive simple extensions of goal-based methods that exhibit some representation learning capability of variational methods, e.g., disentangle underlying factors of variations and automatically determine controllable dimensions, while keeping the learning stability of GCRL.
Viewing variational empowerment as GCRL, we can transfer popular optimization techniques such as relabeling from GCRL to variational empowerment algorithms, and propose latent goal-reaching (LGR) as a more objective, performance-based metric for evaluating the quality of skill (latent goal) discovery. 
We hope that these insights can lay the ground for further developments of more capable and performant algorithms for unsupervised reinforcement learning in future work.

\ifdefined\isaccepted
\section*{Acknowledgements}
The authors are grateful to Igor Mordatch, Lisa Lee, Aleksandra Faust, and Nicolas Heess for helpful discussions and comments.
JC was partly supported by Korea Foundation for Advanced Studies.
\fi

\FloatBarrier

\bibliography{references}  %
\bibliographystyle{icml2021}

\clearpage

\appendix
\onecolumn
\gdef\nohyperref{1}

\icmltitle{Appendix:
    \thepapertitle
}
\medskip

\section{Background: Mutual Information Maximization}
\label{sec:extended-background}

We provide a detailed discussion about mutual information objectives as promised in \Cref{sec:bg_mi}.
\smallskip

\textbf{State-predictive MI}:
Given a generative model of the form $p^\pi(z,s,s')=p(z)\rho^\pi(s|z)p^\pi(s'|s,z)$
where $p^\pi(s'|s,z)=\int \pi(a|s,z) p(s'|s,a) da$, we define the state-predictive MI as,
\begin{align}
    \mathcal{I}(s'; z \mid s)
    &= \mathcal H(s' \mid s) - \mathcal H(s', z \mid s) \\
    &= \mathbb{E}_{(z, s, s') \sim p^\pi(z, s, s')} \left[ \log {p^\pi(s' \mid s, z)} - \log {p^\pi(s' \mid s)} \right]
\end{align}
This is closer to the classic empowerment formulation as in~\citep{klyubin2005all,jung2011empowerment}.
Variational bounds can be derived with respect to actions~\citep{mohamed2015variational,Gregor:ICLRWorkshop2017:VIC} or to future states~\citep{Sharma:ICLR2020:DADS}.
While this objective enables learning state-conditioned skills, we decide to focus on the other variant in this paper.

\textbf{State-marginal MI}:
Similarly, given a generative model of the form $p^\pi(z,s)=p(z)\rho^\pi(s|z)$, the MI can be written as,
\begin{align}
     \mathcal{I}(s; z)  %
     &= \mathcal{H}(z) - \mathcal{H}(z | s) \\
     &= \mathbb{E}_{z \sim p(z)} [ -\log p(z) ] + \mathbb{E}_{z, s \sim p^\pi(z, s)} [ \log p(z \mid s) ] \\
     &= \mathbb{E}_{z \sim p(z), s \sim \pi(z)} [ \log p(z \mid s) - \log p(z) ]   \\
     &\geq \mathbb{E}_{z \sim p(z), s \sim \pi(z)} [ \log q_\lambda (z \mid s) - \log p(z) ]
    \label{eq:var_emp:appendix},
\end{align}
where Eq.~\ref{eq:var_emp:appendix} is a common variational bound for MI~\citep{Barber:NIPS2013:MIbound}
with a variational posterior $q_\lambda (z | s)$ approximating the intractable posterior $p^\pi(z | s)$.
DIAYN~\citep{Eysenbach:ICLR2019:DIAYN} optimizes for this state-marginal MI objective in a entropy-regularized RL setting,
trained with the SAC algorithm~\citep{haarnoja2018soft}.
\medskip
\section{Equivalence between GCRL and Gaussian VGCRL}
\label{sec:full-derivation}

\paragraph{Full Covariance Gaussian.}
The Gaussian discriminator (or the variational posterior) $q_\lambda(z | s)$ should take the following form:
\begin{align}
    q_\lambda(z | s) &= \mathcal N(z; \phi(s), \Sigma(s)) \\
                     &= \frac{1}{\sqrt{(2\pi)^{|\mathcal G|} |\Sigma|}} \exp \left( - \frac{1}{2} (z - \phi(s))^\top \Sigma^{-1} (z - \phi(s)) \right)
\end{align}

\paragraph{Diagonal-Covariance Gaussian.}
If we assume a diagonal covariance $\Sigma(s) = \mathrm{diag}(\sigma^2(s))$,
the discriminator will have the following form:
\begin{align}
    q_\lambda(z | s)
                     &= \frac{1}{\sqrt{(2\pi)^{|\mathcal G|}} \prod_i \sigma_i } \exp \left(
                         - %
                         \sum_i \frac{1}{2 \sigma_i^2}(z_i - \mu_i)^2
                     \right),~~~\text{where}~ \mu_i = [\phi(s)]_i, \sigma_i = [\sigma(s)]_i
    \\
    \log q_\lambda(z | s) &= -|\mathcal G| \log(\sqrt{2 \pi}) + \sum_i - \log (\sigma_i) + \sum_i \left(-\frac{1}{2 \sigma_i^2} (z_i - \mu_i)^2 \right)
    \label{eq:logq-gaussian}
\end{align}
As discussed in \Cref{sec:vge},
the intrinsic reward function for training a goal-conditioned policy for a fixed goal $z$ is given by $r(s) = \log q_\lambda(z | s) - \log p(z)$.

\paragraph{Equivalence to GCRL.}
It is straightforward to see that for a fixed value of $\sigma_i$ (say $\sigma_i = 1.0$), \Cref{eq:logq-gaussian}
further reduces to
\begin{align}
    \log q_\lambda (z | s) = \text{Const} + \sum_i \left( -\frac{1}{2} (z_i - \mu_i)^2 \right)
\end{align}
up to a constant factor.
This can be interpreted as a smooth reward function for reaching a goal $z \in \G$,
or the squared distance $\|z - \mu(s)\|_2^2$ between $\mu(s)$ and $z$ in the goal space $\G$.
A special case of this is when the goal space is set same as the state space ($\mathcal G = \mathcal S$) and a natural identity mapping $\mu(s) = s$ is used,
where the smooth reward function in standard goal-conditioned RL (GCRL) is recovered.
\bigskip
\section{More Experimental Results}
\label{sec:appendix_more_results}
\medskip

In this section, we present additional results for \Cref{sec:experiments-her-lgr}.
\Cref{tbl:summary-key-results} extends \Cref{tbl:summary-key-results-compact}, showing the evaluation metrics for
variants of VGCRL where continuous goal spaces of various dimensions are used.
\Cref{fig:figure-learningcurve-discrete} and \Cref{fig:figure-learningcurve-gaussian} show learning curve plots for
the VGCRL variants with categorical and Gaussian posterior, respectively.

\FloatBarrier

\def\tablefullflag{1}

\def\plotwidth{0.33\textwidth}

\begin{figure}[t]
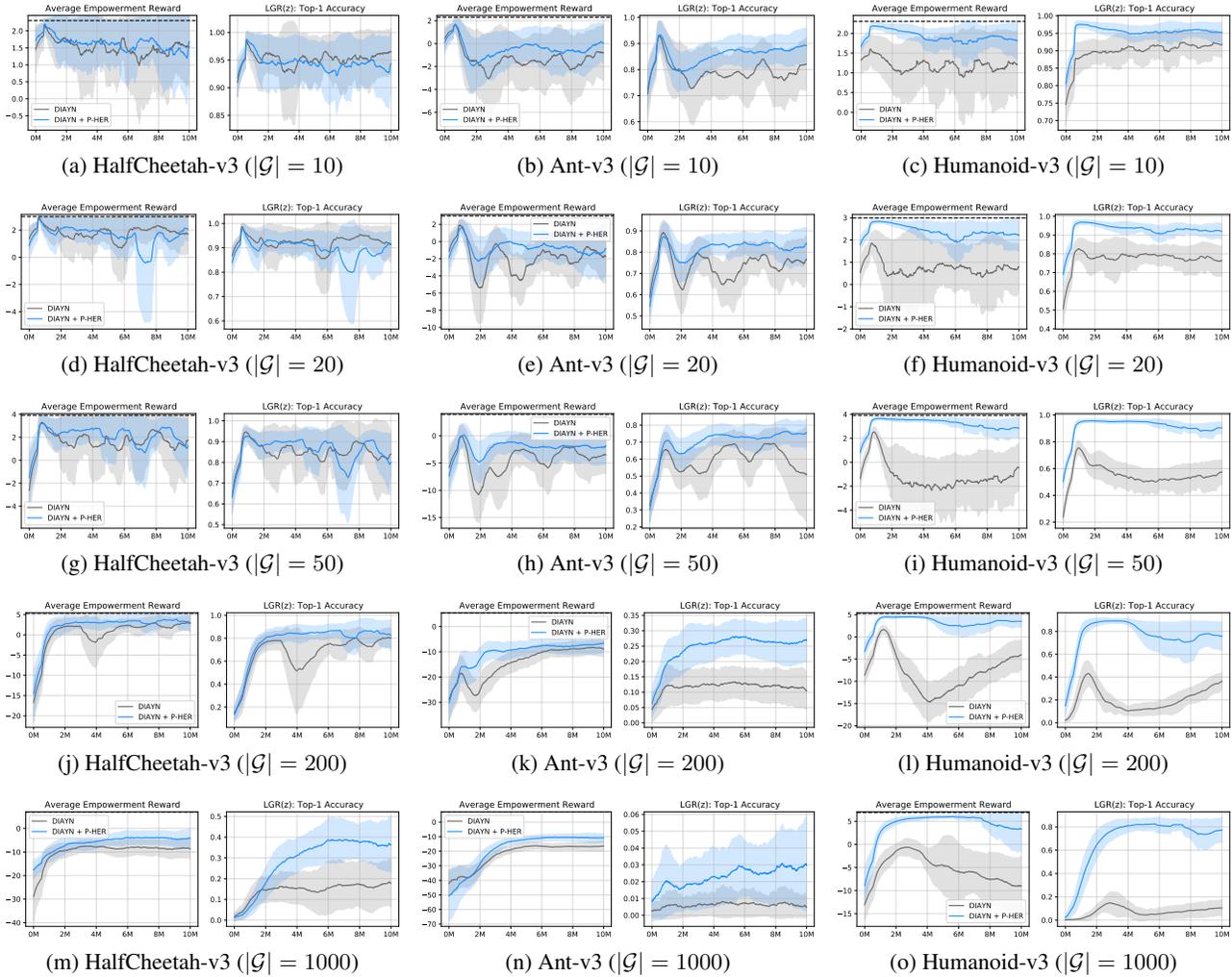

    \vspace*{-20pt}   %
    \captionsetup[subfigure]{aboveskip=+2pt,belowskip=5pt}

\foreach \K in {10,20,50,200,1000}{
    \hfill
    \begin{subfigure}[t]{\plotwidth} %
        \includegraphics[width=\linewidth]{./figures/discrete-her-K\K-halfcheetah-v3.pdf}
        \caption{HalfCheetah-v3 ($|\mathcal G|=\K$)}
    \end{subfigure}
    \hfill
    \begin{subfigure}[t]{\plotwidth} %
        \includegraphics[width=\linewidth]{./figures/discrete-her-K\K-ant-v3.pdf}
        \caption{Ant-v3 ($|\mathcal G|=\K$)}
    \end{subfigure}
    \hfill
    \begin{subfigure}[t]{\plotwidth} %
        \includegraphics[width=\linewidth]{./figures/discrete-her-K\K-humanoid-v3.pdf}
        \caption{Humanoid-v3 ($|\mathcal G|=\K$)}
    \end{subfigure}
}%
    \\%%
    \caption{%
        Extension to \Cref{fig:plot-pher}:
        Learning curves for VGCRL when discrete, categorical goal spaces are used.
        The dashed line denotes the maximum possible reward, achieved when the discriminator $q(z | s)$ is perfect at every time step.
        Overall, we can see P-HER improves the learning process of variational empowerment
        consistently across different environments and the dimensionality of the goal space.
    }
    \label{fig:figure-learningcurve-discrete}
\end{figure}

\def\plotwidth{0.33\textwidth}

\begin{figure}[t]
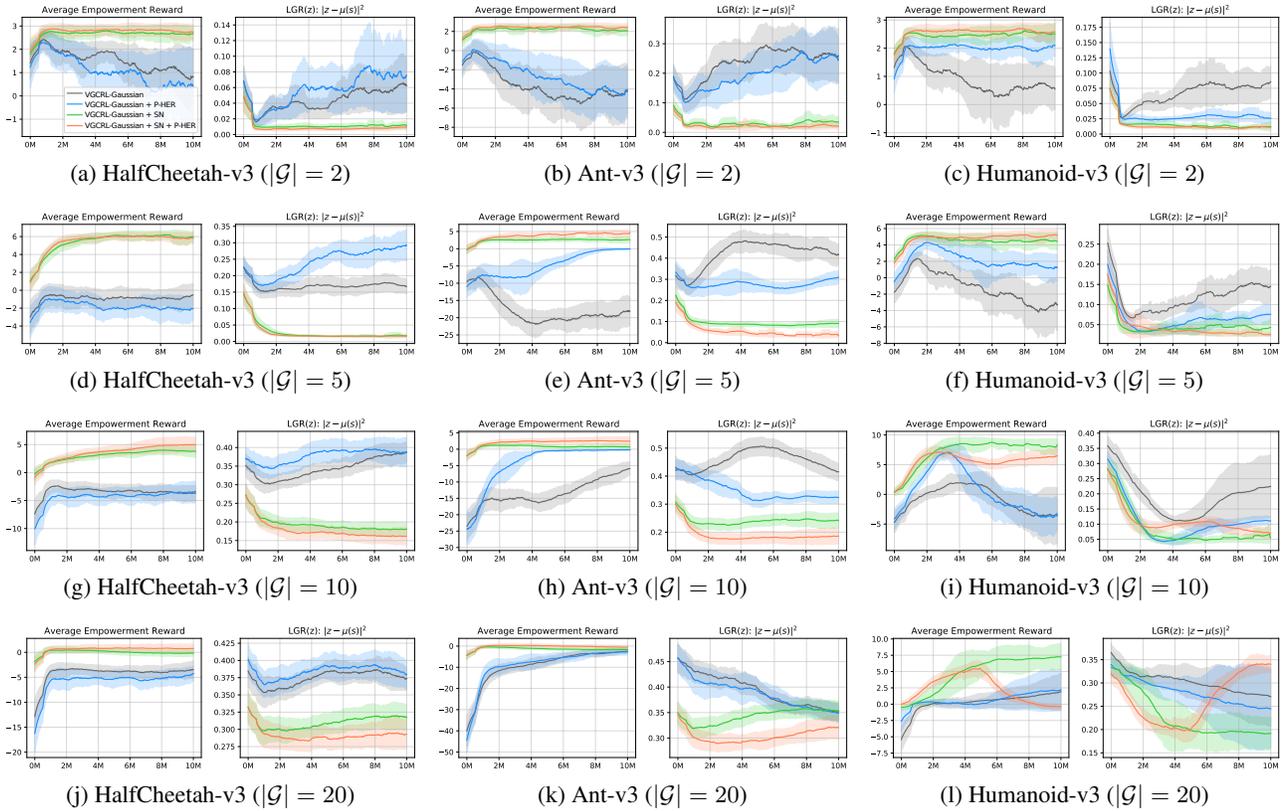

    \vspace*{-20pt}   %
    \captionsetup[subfigure]{aboveskip=+2pt,belowskip=5pt}

\foreach \K in {2,5,10,20}{
    \begin{subfigure}[t]{\plotwidth} %
        \includegraphics[width=\linewidth]{./figures/gaussian-plot-G\K-HalfCheetah-v3.pdf}
        \caption{HalfCheetah-v3 ($|\mathcal G|=\K$)}
    \end{subfigure}
    \hfill
    \begin{subfigure}[t]{\plotwidth} %
        \includegraphics[width=\linewidth]{./figures/gaussian-plot-G\K-Ant-v3.pdf}
        \caption{Ant-v3 ($|\mathcal G|=\K$)}
    \end{subfigure}
    \hfill
    \begin{subfigure}[t]{\plotwidth} %
        \includegraphics[width=\linewidth]{./figures/gaussian-plot-G\K-Humanoid-v3.pdf}
        \caption{Humanoid-v3 ($|\mathcal G|=\K$)}
    \end{subfigure}
}%
    \\%%
    \caption{Extension to \Cref{fig:plot-pher}:
    Learning curves for VGCRL when continuous goal spaces and a family of Gaussian distribution is used for the variational posterior.
    }
    \label{fig:figure-learningcurve-gaussian}
\end{figure}

\clearpage

\vspace*{-5pt}
\section{Details of Environments}
\label{sec:appendix_details}

\vspace*{-5pt}
\subsection{Windy PointMass}

\begin{figure}[h]
    \vspace*{-5pt}
    \centering
    \begin{subfigure}[b]{0.33\linewidth} %
        \includegraphics[width=\linewidth,trim=20 20 20 20,clip]{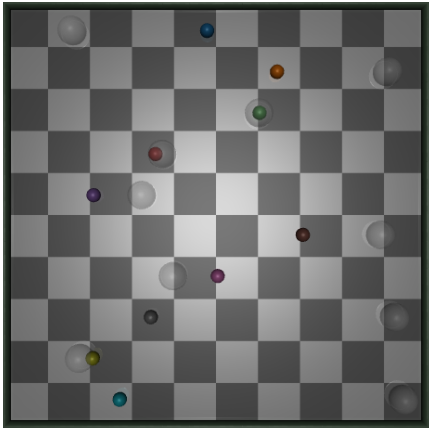}
    \end{subfigure}
    \caption{\small Windy PointMass (10-dimensional).}
    \label{fig:env-pointmass-appendix}
\end{figure}

The (windy) point mass environment is a $N$-dimensional continuous control environment.
The observation space is $2N$-dimensional, each of which describes the position and the velocity per dimension.
Each point mass, one per dimension, can move left and right independently within the arena of range $(-1.5, 1.5)$.
The action space is $N$-dimensional, each of which denoting the amount of velocity acceleration on each dimension.
This generalizes common 2D (planar) point mass environments~\citep{brockman2016gym,tassa2018dmcontrol};
indeed, it is exactly equivalent to the 2D point mass environments when $N = 2$.
The positions of point masses are initialized randomly at each episode.
\Cref{fig:env-pointmass-appendix} shows a target goal location in overlaying transparent spheres
(note that in the experiment we assumed the goal $\mathcal{G}$ to be a $N$-dimensional vector,
same as the observation space) with $\mu(s) = s$.

For the windy point mass used in the experiment, we apply a random external force
sampled from an uniform distribution $U(-R_i, R_i)$ to the point mass on dimension $i$, at every time step.
The range of external force gets higher as the dimension index $i$ increases;
we use a profile of $R_i = 11 \times i$ for $N = 10$
(\ie, $R_0 = 0$ or no force on dimension 0, and $R_9 = 99$ for the last dimension $i=9$) and $[R_0, R_1] = [0, 40]$ for $N = 2$.
With such a large external force, the point mass on dimension $i = 9$ is almost uncontrollable,
mostly bouncing around the external perturbation.
\subsection{Expert State Generation}
\vspace*{-5pt}

To generate target states $s^{1:N}$ in the latent goal reaching metric~\Cref{sec:lgr},
we collected states (observations) randomly sampled from an expert policy's rollout trajectory.
Expert policies are SAC agents successfully trained on the task with multiple target velocities
rather than the standard task (\ie, only moving forward in HalfCheetah, Ant, Humanoid-v3, etc.).
Similar to OpenAI gym's locomotion tasks \citep{brockman2016gym}, we use a custom reward function
$r_x = \mathrm{HuberLoss}(\text{target $x$ velocity} - \text{achieved $y$ velocity})$ and
a similar one for $r_y$ to let the robot move in some directions with the desired target velocities.
The set of target velocities $(v_x, v_y)$ were constructured from the choices of $(-2, -1, -0.5, 0, 0.5, 1, 2)$.
We used the SAC implementation from \citep{TFAgents} with a default hyperparameter setting to train expert policies.
We sample 6 random states from each expert policy, yielding a total of $7^2 \times 6 = 294$ (or $7 \times 6 = 42$ for HalfCheetah) target states
for each environment.
Altogether, this dataset provides a set of states where the agent is posing or moving in diverse direction.

\clearpage
\section{Implementation Details}

For training the goal-conditioned policy, we used Soft Actor-Critic (SAC)~\citep{haarnoja2018soft}
algorithm with the default hyperparameter setting. To represent the discriminator $q(z | s)$
with a neural network, we simply used a 2-layer MLP with (256, 256) hidden units
and ReLU activations.
The heads $\mu(s)$ and $\log \sigma(s)$ are obtained through a linear layer on top of the last hidden layer.
For Gaussian VGCRLs, we employed an uniform prior $p(z) = [-1, 1]^{|\mathcal G|}$ and also applied $\mathrm{tanh}$ bijections
to the variational posterior distribution $q_\lambda(z | s)$ to make the domain of $z$ fit $[-1, 1]^{|\mathcal G|}$.
We also clipped the output of $\log \sigma(s)$ with the clip range $[\log(0.3), \log(10.0)]$ for the sake of numerical stability,
so that the magnitude of posterior evaluations (and hence the reward) does not get too large.

For spectral normalization, we swept hyperparameters $\sigma$ that control the Lipschitz constant
over a range of $[0, 0.5, 0.95, 2.0, 5.0, 7.0]$, and chose a single value $\sigma = 2.0$ that worked best in most cases.
The number of mixtures used in Gaussian Mixture Models is $K = 8$.
The heads $\mu(s)$, $\log \sigma(s)$, mixture weights $\alpha(s)$ are obtained through a linear layer on top of the last hidden layer.

\end{document}